\documentclass[letterpaper, 10 pt, conference]{IEEEtran}  % Comment this line out if you need a4paper
\usepackage{graphicx}
\usepackage{tikz}
\usepackage{comment}
\usepackage{caption}
\usepackage{subfigure}
\usepackage{amsmath,amssymb} % define this before the line numbering.
\usepackage{color}
\usepackage{hyperref}
\usepackage{multicol}
\newcommand\scalemath[2]{\scalebox{#1}{\mbox{\ensuremath{\displaystyle #2}}}}
\newcommand{\mathkomma}{\quad ,}

\newcommand{\snx}[1]{\textit{SalsaNext }{#1}}

\newcommand{\sk}[1]{Semantic-KITTI {#1}}  
\usepackage{booktabs}
\usepackage[accsupp]{axessibility}  % Improves PDF readability for those with disabilities.

\IEEEoverridecommandlockouts                              % This command is only needed if 
\title{\LARGE \bf
Depth- and Semantics-aware Multi-modal  Domain Translation: \\ Generating 3D Panoramic Color Images from  LiDAR Point Clouds*
}

\author{Tiago Cortinhal$^{1}$ and Eren Erdal Aksoy$^{1}$% <-this % stops a space
\thanks{*TThis work was co-funded by the European Union (grant no. 101069576).
Views and opinions expressed are however those of the author(s) only and do
not necessarily reflect those of the European Union or the European Climate,
Infrastructure and Environment Executive Agency (CINEA). Neither the
European Union nor the granting authority can be held responsible for them.}% <-this % stops a space
\thanks{$^{1}$Halmstad University, School of Information Technology,
        Center for Applied Intelligent Systems Research, Halmstad, Sweden
        {\tt\small [tiago.cortinhal,eren.aksoy]@hh.se}}%
}

\makeatletter
\let\@oldmaketitle\@maketitle% Store \@maketitle
\renewcommand{\@maketitle}{\@oldmaketitle% Update \@maketitle to insert...
\centering 
  \includegraphics[width=0.8\linewidth, height=0.5\linewidth]
   {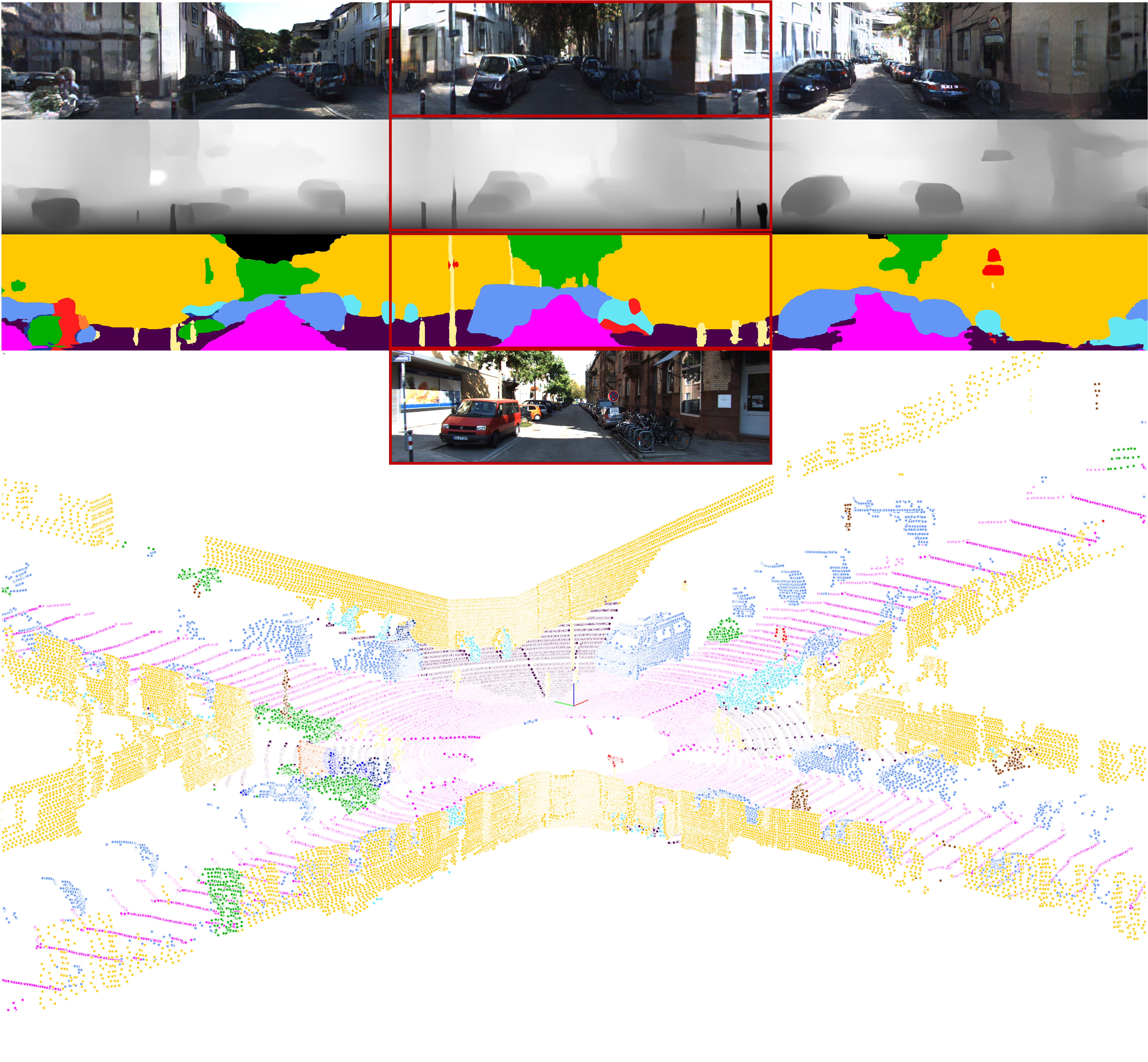}
   \captionof{figure}{
   We propose a modular generative neural network framework that receives a full 3D LiDAR point cloud and returns a panoramic colour image enriched with depth information by solely relying on the semantics of the scene. The framework first applies semantic segmentation to the raw LiDAR scan (the bottom image). Next, our proposed model TITAN-Next translates the LiDAR segments to the camera semantic segments with the corresponding depth information (the second and third images from the top), which are then converted back to the panoramic colour images (first image) by an additional generative model.  The red frame indicates the region that the ground-truth camera image (the fourth image) corresponds to.
   }\bigskip
    \label{fig:introSample} 
   }% ... an image

\begin{document}

\maketitle
\thispagestyle{empty}
\pagestyle{empty}

%%%%%%%%%%%%%%%%%%%%%%%%%%%%%%%%%%%%%%%%%%%%%%%%%%%%%%%%%%%%%%%%%%%%%%%%%%%%%%%%
\begin{abstract}
This work presents a new depth- and semantics-aware conditional generative model, named TITAN-Next, for cross-domain image-to-image translation in a multi-modal setup between LiDAR and camera sensors. The proposed model leverages scene semantics as a mid-level representation and is able to translate raw LiDAR point clouds to panoramic RGB-D camera images by solely relying on semantic scene segments. We claim that this is the first framework of its kind, and it has practical applications in autonomous vehicles, such as providing a fail-safe mechanism and augmenting available data in the target image domain. The proposed model is evaluated on the large-scale and challenging Semantic-KITTI dataset, and experimental findings show that it considerably outperforms the original TITAN-Net and other strong baselines by 23.7$\%$ margin in terms of IoU.
\end{abstract}

%%%%%%%%%%%%%%%%%%%%%%%%%%%%%%%%%%%%%%%%%%%%%%%%%%%%%%%%%%%%%%%%%%%%%%%%%%%%%%%%
\section{INTRODUCTION}

There are several attempts at converting images from one domain to another type-similar domain, known as domain translation, including mapping sketches  or segmentation masks to RGB images~\cite{pix2pix2017,CycleGAN2017,Wang2017}. 
These cross-domain \textit{image-to-image} translations  have recently received significant attention in computer vision and robotics. 
However, the cross-domain mapping in a multi-modal setup, for instance, between LiDAR and camera sensors, is a challenge and still remains in its infancy~\cite{Milz2019Points2Pix3P,Kim_2019,Kim2020ColorIG,Shalom2021}. 
Primarily,the challenge with LiDAR scans stems from their inability to capture the rich texture details that RGB images provide, resulting in 3D point clouds that are sparse and non-uniformly dispersed, unlike the more structured 2D images from RGB sensors.
This, as a result, makes the image synthesis  from raw point clouds non-trivial. 
In this work, we address the challenge of \textit{multi-modal domain translation} by leveraging the scene semantics as a mid-level  representation between different sensor modalities with  unique data formats.
To overcome this challenge, we introduce a new depth- and semantics-aware  conditional generative model, named TITAN-Next (generaTive domaIn TrANslation Network), which adversarially learns to translate LiDAR point clouds to RGB-D camera images by solely relying on the semantic scene segments (see Fig.~\ref{fig:introSample}).

TITAN-Next is built upon the model TITAN-Net~\cite{Cortinhal_2021_ICCV}\footnote{To be removed in the final version: The TITAN-Net framework, which is the early version of the here proposed model TITAN-Next,  received the \textit{``best student paper award"}  at the \href{https://www.ai4ad.net/}{IJCAI 2021 AI4AD workshop.}}, which is embedded in a modular generative neural network framework.
We, here, revise this modular framework (see Fig.~\ref{fig:pipeline})  to make the original TITAN-Net concurrently predict  the depth information while generating  RGB camera images with higher fidelity.
More specifically, we have the following contributions:
\begin{itemize}

\item We introduce a multi-scale \textit{feature pyramid module} in the original TITAN-Net  encoder and decoder.
 To enforce stability and improve learning at different scales, we inject a downsampled version of the LiDAR input into each level of the encoder. Similarly, we extract the features from each decoder block, upsample, and concatenate them with the final output. % before merging them and reconstructing a more detailed semantic segmentation map. 
This  multi-scale feature aggregation   yields a significant improvement in robustness, accuracy, and  convergence speed.  

\item We further append a lightweight \textit{depth head} to the TITAN-Net decoder to enrich the generated scene image with  the depth information derived directly from the LiDAR input. This eventually leads to generating RGB-D images from given point clouds. 

\item We  also report comprehensive quantitative and qualitative evaluations using  the large-scale, challenging  \sk dataset~\cite{semantickitti}. 
Our experimental findings show that our new framework considerably outperforms the original TITAN-Net and other strong baselines by a large margin. 
Our code and trained models are available at \url{https://github.com/TiagoCortinhal/TITAN-Next} to encourage further research on
the subject.

\end{itemize}

To the best of our knowledge, the here proposed \textit{multi-modal domain translation} framework is the first of its kind that can  generate photo-realistic RGB-D images from raw 3D  point clouds by solely relying on the sensor agnostic scene semantics. 
This has practical applications in autonomous vehicles.  
For instance, an abrupt failure of RGB and depth sensors can be immediately    substituted by the synthesized RGB-D image streams to boost the subsequent sensor fusion  and planning modules. 
This   introduces a natural fail-safe mechanism to the vehicle to prevent  a system collapse. 
Another potential application is augmenting the available data in the target image domain. The generated semantic scene segments can be employed to fabricate vast variants of the original scene by feeding  data from different distributions without any additional effort. 
Furthermore, the  framework can also be applied to colourize point clouds without relying on real RGB camera readings. 

 \section{RELATED WORK}
There exists a vast number of work  on generating novel scene images~\cite{wang2018vid2vid, nvidia_vid2vid, surfelgan} and rendering   point clouds~\cite{Ouyang2017ACS, Milz2019Points2Pix3P, AtienzaCVPR2019, Kim_2019, Kim2020ColorIG, SongECCV2020}. Coupling  both addresses the multi-modal domain translation, which still remains an unsolved challenging problem.   

There is a large corpus of work in image-to-image translation, such as converting an image from one domain (e.g., from a segmentation space) to a different one (e.g., to an RGB space)~\cite{pix2pix2017,CycleGAN2017}. 
Image semantics has also been injected to accelerate such a domain transfer process~\cite{slmarchildon2021,Roy2019,Tomei_2019_CVPR,mo2018}. 
Similarly, photo-realistic {video-to-video} translations, i.e., converting image sequences by considering temporal cues, have  been intensively studied~\cite{wang2018vid2vid,nvidia_vid2vid}. 
  
Recent works have, so far, focused on bridging the gap between the same sensor modalities, i.e., either between RGB camera images~\cite{2018arXiv181112833V,2020arXiv200405498Y,CycleGAN2017} or between LiDAR sensor readings~\cite{9341508,8814047}.
The literature is still lacking of translation across different sensor modalities with distinct   characteristics, which is  of particular interest  in this work.

Most recent works on translating  3D  point clouds to  2D  RGB images are based  on conditional Generative Adversarial Networks (cGANs). This generative process relies on a conditional constraint, ranging from a one-hot vector of a given class set to any other type of data that conditions the generative process. For instance, works like~\cite{Ouyang2017ACS} exploit the informativeness of LiDAR point clouds projected onto the RGB space to render a depiction of the RGB image, where the conditions   for the cGAN are real images. 
Following a similar approach, predefined  background image patches are employed in Milz et al.~\cite{Milz2019Points2Pix3P} to bias a GAN model while exploiting the 3D representation in the bottleneck.
The recent method \textit{pc2pix} introduced in~\cite{AtienzaCVPR2019} relies on auxiliary cGAN to render a point cloud to an RGB image from the desired camera angle without employing any surface reconstruction operation. This way, \textit{pc2pix} can generate different images by applying arithmetic operations in the latent space.

As an alternative to cGAN-based adversarial learning approaches, asymmetric encoder-decoder networks~\cite{Kim_2019} and U-NET models~\cite{Kim2020ColorIG} are also introduced to render point clouds in a rather supervised fashion.
Likewise, the work in~\cite{SongECCV2020} introduces an end-to-end framework with a point cloud encoder and an RGB decoder enriched with a refinement network in order to enhance  the image quality generated from novel viewpoints in a point cloud.

The works from~\cite{Milz2019Points2Pix3P,Kim_2019,Kim2020ColorIG,Shalom2021} can be considered the most similar to our method. Although these methods can, to some degree, generate scene images from a given point cloud, they  cannot process  unstructured and sparse  \textit{full-scan} LiDAR point clouds due to computational constraints. %lack of enough capacity. 
Furthermore, these approaches neither incorporate scene semantics nor predict the depth map for the generated image.
%also completely  omit the depth information and scene semantics. 
In contrast,  during inference, we utilize the complete LiDAR scan to generate a panoramic scene image, and the corresponding depth information. Conversely, during training, we work with cropped sections of the LiDAR data aligned with the camera's field of view.

Regarding semantic scene segmentation, which is a well-studied topic, we have a particular interest in projection-based models~\cite{salsanext,milioto2019iros,salsanet} to segment LiDAR point clouds. This is because the spherical projection, for instance, yields  the native 2D view of the LiDAR scan, which allows for generating panoramic RGB counterparts.
There is also a large corpus of work on the semantic segmentation of camera images~\cite{2020arXiv200402307M,2020arXiv200311883Z} with a surge of new papers focusing on Vision Transformer approaches~\cite{2020arXiv200510821T,2021arXiv211109883L}. 
Thanks to the modular structure of our framework, any of these  state-of-the-art segmentation models can be seamlessly integrated into the proposed pipeline. 

\setcounter{figure}{1}
\begin{figure*}
    \centering
    \includegraphics[width=1\linewidth]{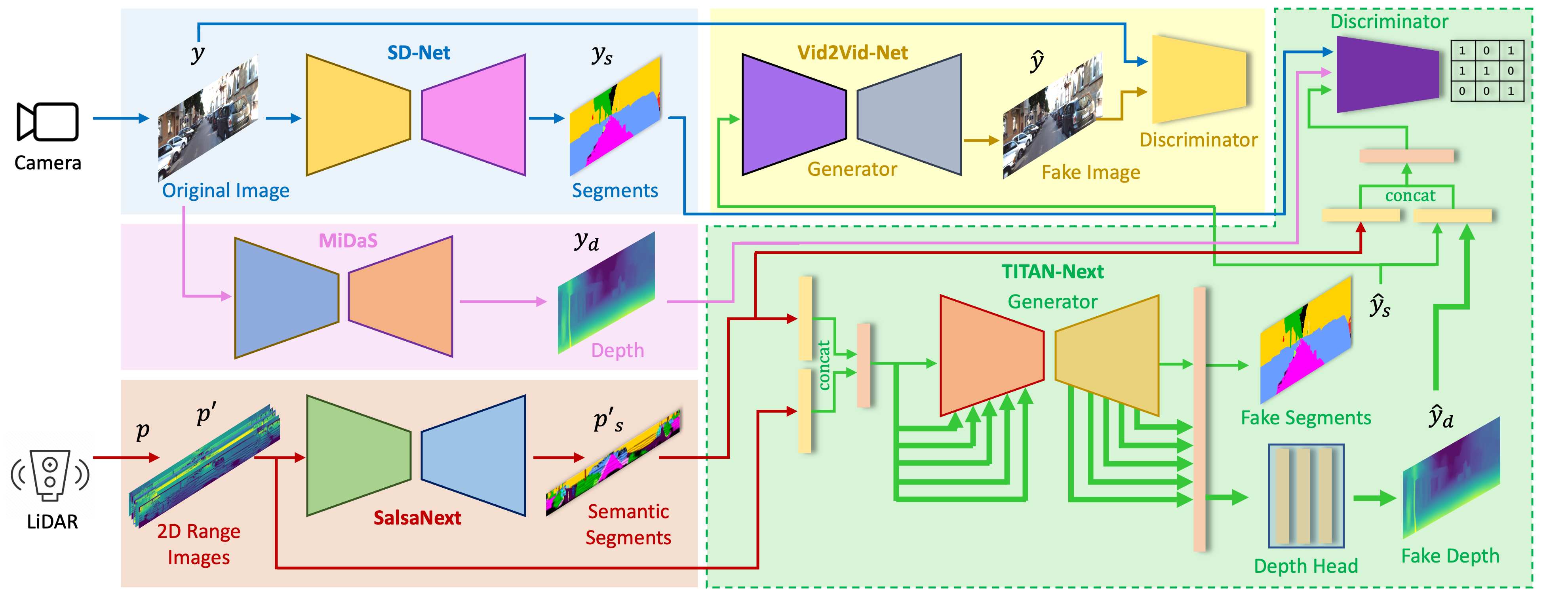}
\caption{%
   Our proposed modular framework has five neural networks, each of which is depicted by a unique background colour. Updates introduced by TITAN-Next are indicated using bold arrows, marking the advancements over the TITAN-Net\cite{Cortinhal_2021_ICCV} framework.
   In the red box, a 3D LiDAR point cloud $p$ is first projected onto the 2D range image plane $p^{\prime}$ to be segmented by \textit{SalsaNext}~\cite{salsanext}, predicting $p^{\prime}_{s}$.
   Likewise, the corresponding RGB camera image $y$ is processed by SD-Net~\cite{2020arXiv200510821T}  to predict semantic segments $y_s$, as depicted by the blue box. 
   The depth information $y_d$ is meanwhile estimated by   MiDaS~\cite{Ranftl2020}  as shown in the purple box.
   The green box highlights the  TITAN-Next model, where the \textit{Generator}  is conditioned on the concatenated $p^{\prime}_{s}$ and $p^{\prime}$ to generate the fake camera segment map $\widehat{y}_{s}$ and the fake depth information $\widehat{y}_{d}$. 
   The TITAN-Next \textit{Discriminator} is also conditioned on  $p^{\prime}_{s}$ while comparing $\widehat{y}_{s}$ and $\widehat{y}_{d}$ with the expected $y_s$ and  $y_d$, respectively. 
   Finally, as shown in the yellow box, the generated fake camera segment  $\widehat{y}_{s}$ is processed by Vid2Vid-Net~\cite{wang2018vid2vid} to  synthesize  the realistic RGB image  $\widehat{y}$.
   Note that the blue and purple boxes are employed during training only. 
   }
    \label{fig:pipeline}
\end{figure*}
%%%%%%%%%%%%%%%%%%%%%%%%%%%%%%%%%%%%%%%%%%%%%%%%%%%%%%%%%%%%%%%%%%%%%%%%%%%%%%%%

\section{Method}

As depicted in Fig.~\ref{fig:pipeline}, we propose a modular generative framework, which involves five individually trained neural networks to translate a given LiDAR point cloud to the corresponding RGB-D image. 
TITAN-Next shown in the green box in Fig.~\ref{fig:pipeline} constitutes the main contribution of this work. 
Updates introduced by TITAN-Next are indicated using bold arrows, marking the advancements over the TITAN-Net\cite{Cortinhal_2021_ICCV} framework.
We follow a cGAN approach, where our model TITAN-Next is conditioned on  the semantic segment maps of the spherically projected LiDAR point clouds. 
%TITAN-Next maps the segments of the LiDAR data to that of the image space to generate the  RGB counterparts with the  corresponding depth information. 
%
More specifically, given a LiDAR point cloud $p \in \mathbb{R}^{n \times 4}$, where $n$ is the cardinality of the  cloud storing  $x$, $y$, $z$ point coordinates and $i$ intensity values, TITAN-Next not only learns the mapping between the LiDAR and camera image semantic segments but also estimates the corresponding depth information.
Estimated image segments and depth cues are further processed in order to generate the corresponding four channel RGB-D image of the scene $y \in \mathbb{R}^{w \times h \times 4}$, where $w$ and $h$ are the image width and height.
%
%Consequently, our proposed domain translation can be formulated  as $G_{\{p,s\} \mapsto y}: \{p,s\} \mapsto y$, which is conditioned on the semantic segment maps $s$.
%
In the following subsections, we will detail the framework and improvements.

%%%%%%%%%%%%%%%%%%%%%%%%%%%%%%%%%%%%%%%%%%%%%%%%%%%%%%%%%%%%%%%%%%%%%%%%%%%%%%%%
\subsection{LiDAR Point Cloud Segmentation}
Our framework starts with the semantic segmentation of LiDAR point clouds, as shown in the red box in Fig.~\ref{fig:pipeline}.

Following the works in~\cite{salsanext,milioto2019iros}, we apply  a spherical projection to a given point cloud  $p \in \mathbb{R}^{n \times 4}$ by mapping each point in $p$ into an image coordinate. 
The final output of this projection is $p^{\prime} \in \mathbb{R}^{\mathit{w^{\prime}} \times \mathit{h^{\prime}} \times 5}$, i.e., an image with the size of $\mathit{w^{\prime}} \times \mathit{h^{\prime}}$. 
This way, the unstructured point cloud $p$ is converted into  a structured image with five channels representing  $(x,y,z)$ coordinates, intensity, and range values. 
 
Note that the projection of $p$  to $p^{\prime}$ helps to solve the  correspondence problem between the two distinct sensor data readings without leading to an enormous information loss since the spherical projection is the native 2D representation of a 3D LiDAR point cloud and reduces issues like shadows and distortions obtained in other projection types such as top-view~\cite{salsanet}.

Next, the projected LiDAR data $\mathit{p^{\prime}}$ is fed to an off-the-shelf   segmentation network \textit{SalsaNext}~\cite{salsanext} to predict point-wise semantic    labels  in a 2D image format, $p^{\prime}_{s} \in \mathbb{R}^{\mathit{w^{\prime}} \times \mathit{h^{\prime}}}$.

%%%%%%%%%%%%%%%%%%%%%%%%%%%%%%%%%%%%%%%%%%%%%%%%%%%%%%%%%%%%%%%%%%%%%%%%%%%%%%%%
\subsection{Camera Image Segmentation}
As depicted in the blue box in Fig.~\ref{fig:pipeline}, we also semantically segment the corresponding RGB camera images of the scene, synchronized with the LiDAR data. 
Note that this segmentation process is only called during training.

In this step,  a state-of-the-art  segmentation network SD-Net~\cite{2020arXiv200510821T} is employed to convert the original RGB images $\mathit{y}$ to  one-channel segment  masks,  $y_{s} \in \mathbb{R}^{\mathit{w} \times \mathit{h}}$, where $w$ and $h$ are the width and height of the input RGB image $\mathit{y}$.

%%%%%%%%%%%%%%%%%%%%%%%%%%%%%%%%%%%%%%%%%%%%%%%%%%%%%%%%%%%%%%%%%%%%%%%%%%%%%%%%
\subsection{Camera Image Depth Estimation}
Our  pipeline also incorporates the depth information of the scene image, as shown in the purple box in Fig.~\ref{fig:pipeline}.

The primary role of the depth information in our pipeline is not only to boost the translation from LiDAR semantic segments to camera counterparts, but also to estimate the depth cues for the generated final scene images. For this purpose, we employ MiDaS~\cite{Ranftl2020} as an off-the-shelf  depth estimator to produce one-channel depth cues,  $y_{d} \in \mathbb{R}^{\mathit{w} \times \mathit{h}}$, where $w$ and $h$ are the width and height of the input RGB image $\mathit{y}$.
We also note that this depth estimation step is called during training only.

%One lacking aspect of TITAN-Net is that it does not exploit the LiDAR scan properly; the depth cues from the LiDAR can be helpful to not only predict the semantic segments but extend them to predict the depth information— allowing for a possible generation of synthetic LiDAR scans. Regardless, this work focuses on the Domain adaptation from LiDAR to Camera space.
%We use MiDaS~\cite{Ranftl2020} as an out-of-the-self depth estimator to produce our ground truth for modelling the depth information. MiDaS is a combination of a ResNet~\cite{7780459} feature extractor and a Vision Transformer for Dense Prediction (DPT)~\cite{Ranftl2020}. The novelty in MiDaS is the combination of several depth databases in an attempt at Zero-Shot transfer. 

%%%%%%%%%%%%%%%%%%%%%%%%%%%%%%%%%%%%%%%%%%%%%%%%%%%%%%%%%%%%%%%%%%%%%%%%%%%%%%%%
\subsection{From LiDAR to RGB segments and Depth}
\label{section:segmentation}
The translation from LiDAR point clouds to the RGB-D space starts once the semantic segments of both domains and the image depth information are available, i.e., $p'_s$, $y_s$, and $y_d$, as highlighted by the green box in Fig.~\ref{fig:pipeline}.
We here note that this green box represents our core contribution in this work, i.e., a new depth- and semantics-aware cGAN model, named TITAN-Next.

\textbf{TITAN-Next:} is built upon the base TITAN-Net model~\cite{Cortinhal_2021_ICCV} and has two core novel contributions, yielding  depth-aware and more accurate transfer of LiDAR semantic segments to the image space. These novelties are the integration of the \textit{feature pyramid module} in the encoder and decoder, and introducing a new \textit{depth branch} to the  TITAN-Net generator as described below.  

\textit{\textbf{Feature Pyramid Module}} introduces multiscale feature aggregation to the original TITAN-Net encoder (see the green box in Fig.~\ref{fig:pipeline}). 
We feed the concatenated  LiDAR  input  $p'$ and $p'_s$ into each encoder layer to enforce meaningful bottleneck representations. 
Following the work of~\cite{2016arXiv161203144L}, we hypothesize that injecting  downsampled versions of the input at each level would force the encoder to extract more descriptive and more robust feature representations. This could be seen as one level of DenseNet~\cite{2016arXiv160806993H}, where only the input is densely propagated in the encoder. 
%Note that the same treatment is not applied to the decoder, as we want the decoder to be more dedicated to the target domain, i.e., the RGB space.
In the decoder, we extract the representation of each encoder layer, upsample them accordingly to reach the original RGB input dimension, and finally concatenate them before the final prediction layer. 

The final output is  composed of a merged version of each decoder level. 
The main idea behind using each intermediary representation in the final prediction is to boost the quality of the predicted segments for the less frequent and smaller classes (e.g., traffic signs and poles) by ensuring that the representations are expressive.

\textit{\textbf{Depth Head}} is  attached to the decoder  as a new branch  to concurrently predict the depth information  $\widehat{y}_{d}$ while generating the image segment maps  $\widehat{y}_{s}$. 
The depth head consists of 3 standard convolutional layers and receives the features of the last decoder layer as an input.
 
We here highlight the fact that TITAN-Next generator employs the scene depth information ($y_d$ in~Fig.~\ref{fig:pipeline}) neither as an input nor as a condition. 
Due to our new \textit{Feature Pyramid Module} introduced above, TITAN-Next can concurrently generate both the depth map and camera image segments by  relying only  on the LiDAR data ($p'$ and $p'_s$). 
This new contribution makes TITAN-Next independent of the camera depth cues during inference. 
Recall that the purple box   in~Fig.~\ref{fig:pipeline} is called by the TITAN-Next discriminator during training only. 

In contrast to the original TITAN-Net, the TITAN-Next discriminator has a new task of jointly distinguishing between semantic segmentation ($\widehat{y}_{s}$) and depth estimation ($\widehat{y}_{d}$) instead of having a separate discriminator for each output type.
The main reason is that both outputs share enough similarities about the scene context, which can help the discriminator learn the joint distribution. 

%Following the works of~\cite{CycleGAN2017, 2016DeepMultiScale} we omit the noise vector $z$. Our input to the Generator of TITAN-Next will comprise the spherical representation of the LiDAR and its semantic segmentation map. The latter being our condition and the former our input.

%%%%%%%%%%%%%%%%%%%%%%%%%%%%%%%%%%%%%%%%%%%%%%%%%%%%%%%%%%%%%%%%%%%%%%%%%%%%%%%%

\textbf{TITAN-Next Loss Function:} 
In contrast to TITAN-Net~\cite{Cortinhal_2021_ICCV}, we here use an additional Weighted Cross-Entropy while generating the segmentation masks ($\widehat{y}_{s}$) and employ three different loss components for the depth estimation ($\widehat{y}_{d}$). 

To generate more accurate segmentation masks, particularly for the underrepresented classes such as bicycles and traffic signs, we weight the softmax cross-entropy loss with the inverse square root of the class frequency as described in~\cite{salsanext}. 
The overall   segmentation loss is $\mathcal{L}_{seg} =  \mathcal{L}_{wce} + \mathcal{L}_{ls}$. 

$\mathcal{L}_{wce}$ is the Weighted Cross Entropy loss  defined as: 
\begin{equation}
\scalemath{0.85}{
\mathcal{L}_{wce}(y,\hat{y}_{s_i}) = -\sum_{i} \alpha_{i}p(y_{s_i})log(p(\hat{y}_{s_i}) ~,~~ with ~~~
 \alpha_{i} = 1/\sqrt{f_{i}}    \mathkomma
}
\end{equation}

where ${y_s}_i$ and $\hat{y}_{s_i}$ are the true and predicted class labels and $f_{i}$ stands for the frequency of appearance in the dataset. 

$\mathcal{L}_{ls}$ is the Lovász-Softmax loss~\cite{berman2018lovasz}   described as:
\begin{equation}
\scalemath{0.77}{
\mathcal{L}_{ls} = \frac{1}{|C|}\sum_{c\in C} \overline{\Delta_{J_c}}(m(c)) ~,~~ and ~~~
m_i(c) = \left\{
\begin{array}{l l}
  1-x_{s_i}(c) &   \text{if ~ $c = y_{s_i}(c)$    } \\
  x_{s_i}(c) &   \text{otherwise}\\
\end{array}
\right.
~,
}
\end{equation}
 
where $|C|$ represents the class number, $\overline{\Delta_{J_c}}$ defines the Lov\'{a}sz extension of the Jaccard index, $x_i(c) \in [0,1]$ and $y_i(c) \in \{-1,1\}$ hold the predicted probability and ground truth label of the pixel $i$ for class $c$, respectively.
 
When it comes to the depth estimation, TITAN-Next combines  three losses:
$\mathcal{L}_{depth} = \mathcal{L}_{ids} + \mathcal{L}_{mae} + \mathcal{L}_{ssim}.$

Inverse Depth Smoothness loss, $\mathcal{L}_{ids}$, is first introduced in~\cite{2017arXiv171200175W}. The main idea behind this loss term is to enforce flat slopes between different objects at different depth levels. A smoothness cost is implemented using the Laplacian of the image intensity, where higher intensity means that the pixel is most likely on an edge or at a corner. In such regions, a lower cost is imposed:
\begin{equation}
    \mathcal{L}_{ids}(y_{d_i}) = e^{-\nabla^{2}\mathcal{I}(\hat{y}_{d_i})(|\partial_{xx}y_{d_i}| + | \partial_{xy}y_{d_i}| + |\partial_{yy}y_{d_i}|)} ~,
\end{equation}
where $\nabla^{2} \mathcal{I}(x_i)$ represents the Laplacian of the image intensity, and
$\partial$ the partial derivatives of the depth estimation. 

Mean Absolute Error (MAE)  loss, $\mathcal{L}_{mae}$ is computed as:
\begin{equation}
    \mathcal{L}_{mae} = \frac{1}{N} \sum_{i \in N} | y_{d_i} - \hat{y}_{d_i} | ~,
\end{equation}
where $N$ is the total number of pixels in the image, $y_i$ is the ground-truth, and $\hat{y}_i$ is the network prediction.

Structural Similarity Index Measure (SSIM)~\cite{fid} measures the similarity between two images regarding their luminance, contrast, structure, and covariance matrix.
The loss form of this metric, $\mathcal{L}_{ssim}$, can be computed as:
\begin{equation}
    \mathcal{L}_{ssim}(y_{d},\hat{y}_{d}) = \frac{1-SSIM(y_{d},\hat{y}_{d})}{2} ~,
\end{equation}
where $y_d$ and $\hat{y}_d$ represent the ground truth   and prediction, respectively.

In addition to these supervised losses, a Wasserstein Gradient Penalty~\cite{2017arXiv170400028G} is used for the adversarial training. The loss can be described as:
\begin{equation}
\begin{aligned}
\mathcal{L}_{D}^{WGANGP} = \mathbb{E}_{(\tilde{x},\tilde{y})\sim \mathbb{P}_g}[D(\tilde{x},\tilde{y})] - \mathbb{E}_{(x,y)\sim \mathbb{P}_r}[D(x,y)] + \\ 
\lambda \mathbb{E}_{(\hat{x},\hat{y})\sim \mathbb{P}_{(\hat{x},\hat{y})}}[(|| \nabla_{(\hat{x},\hat{y})} D((\hat{x},\hat{y}))||_2 -1)^2] ~\mathkomma
\end{aligned}
\end{equation}
\begin{equation}
\begin{aligned}
&\qquad    \mathcal{L}_{G}^{WGANGP} = - \mathbb{E}_{(\tilde{x},\tilde{y}) \sim \mathbb{P}_g}[D((\tilde{x},\tilde{y}))] ~,
\end{aligned}
\end{equation}
where $\mathcal{P}_r$,$\mathcal{P}_g$,$\mathcal{P}_{\hat{x}}$ represent the real, generated, and sampling probabilities  and $x$ defines the segments and $y$ is the depth information. 
The sampling probabilities are uniformly sampled along straight lines between $\mathcal{P}_r$ and $\mathcal{P}_g$ as in~\cite{2017arXiv170400028G} and correspond to the gradient penalty, $\lambda$ is set to 10 following the original implementation.

The  final TITAN-Next loss is the linear combination of the above losses, $\mathcal{L}= \mathcal{L}_{seg}+\mathcal{L}_{depth}+\mathcal{L}_{wgan\_gp}$.

%%%%%%%%%%%%%%%%%%%%%%%%%%%%%%%%%%%%%%%%%%%%%%%%%%%%%%%%%%%%%%%%%%%%%%%%%%%%%%%%
\subsection{From  RGB segments and Depth to RGB-D Images}
Finally, the generated  camera segments  $\widehat{y}_{s}$   are converted to RGB scene images, $\widehat{y}$, as highlighted in the yellow box in Fig.~\ref{fig:pipeline}. 
This translation of $\widehat{y}_{s}$ to $\widehat{y}$ is achieved by another   off-the-shelf state-of-the-art cGAN model Vid2Vid-Net~\cite{wang2018vid2vid}. 
By combining $\widehat{y}$ with the predicted depth maps $\widehat{y}_{d}$, we then generate the RGB-D scene image, which is the final output of our modular pipeline in Fig.~\ref{fig:pipeline}.

%%%%%%%%%%%%%%%%%%%%%%%%%%%%%%%%%%%%%%%%%%%%%%%%%%%%%%%%%%%%%%%%%%%%%%%%%%%%%%%%
\section{Experimental Setup}
%%%%%%%%%%%%%%%%%%%%%%%%%%%%%%%%%%%%%%%%%%%%%%%%%%%%%%%%%%%%%%%%%%%%%%%%%%%%%%%%

\subsection{Implementation Details}
For a fair comparison, the same training protocols introduced~\cite{Cortinhal_2021_ICCV} are followed here. 
We use Adam~\cite{adam2015} with a learning rate of $1\times 10^{-4}$ and $(0.5,0.999)$ as $(\beta_1,\beta_2)$. 
The batch size is fixed at 6, while the dropout probability is set to 0.2 throughout the network.
The spherical projection size is fixed at $512\times 64$ pixels, being the corresponding area of the RGB space. 
The camera segmentation output is fixed at $1241\times 376$ pixels.

We also performed data augmentation by flipping randomly around the $y$-axis and by randomly dropping points before creating the projection. Both data augmentation techniques are set at $0.5$ probability. 

For more details about the implementation, we refer to the publicly available source code~\url{https://github.com/TiagoCortinhal/TITAN-Next}.
%%%%%%%%%%%%%%%%%%%%%%%%%%%%%%%%%%%%%%%%%%%%%%%%%%%%%%%%%%%%%%%%%%%%%%%%%%%%%%%%
\subsection{Dataset}
We use the large-scale challenging  \sk dataset~\cite{semantickitti} for the experimental evaluation.
\sk  has over 43K point-wise annotated full LiDAR scans with 19 distinct classes.
We follow the same training, validation, and test splits used in~\cite{Cortinhal_2021_ICCV}.

Since \sk does not provide annotated camera images, we exploit the semantic segmentation labelling of the Cityscapes dataset by inferring pseudo-labels via SD-Net~\cite{2019arXiv190710659O}. We follow the same mapping between Cityscapes and \sk as in~\cite{Cortinhal_2021_ICCV}. Note that due to this mapping between the two datasets, the total class number is reduced to 14 as described in~\cite{Cortinhal_2021_ICCV}.
 
%%%%%%%%%%%%%%%%%%%%%%%%%%%%%%%%%%%%%%%%%%%%%%%%%%%%%%%%%%%%%%%%%%%%%%%%%%%%%%%%

\subsection{Evaluation Metrics}
\label{sec:metrics}
The following metrics have been used to evaluate the quality of the generated segmentation masks, depth maps, and rendered images.
%Note that $T$ refers to the total number of pixels, $y$ is the ground truth and $\hat{y}$ represents the  prediction.

\textbf{Intersection over Union (IoU):} measures the overlap between the real and predicted segments and is primarily used for evaluating the accuracy of the generated semantic segmentation maps. The higher the IoU score, the better. 

\textbf{Fr\'echet Inception Distance (FID):} measures the visual quality of the generated images and is sensitive to artifacts and noise~\cite{fid}. A low FID score indicates that the generated image is closer to the expected real image. 

\textbf{Sliced Wasserstein Distance (SWD):}  compares  structural similarities between the generated   and real images at different scales~\cite{karras2018progressive} by computing a Laplacian Pyramid~\cite{pyramid} starting with a resolution of $16\times16$ that is doubled and repeated until reaching the full image size. A lower SWD value represents better image quality.

\textbf{Absolute Relative Error (AbsRel):} represents the magnitude of the differences between ground-truth and prediction. The lower, the better. 

%It can be formulated as:
%\begin{equation}
%   \frac{1}{|T|}\sum_{y \in T} \frac{|y-\hat{y}|}{\hat{y}}   
%\end{equation}

\textbf{Squared Relative Difference (SqRel):} is the squared difference between ground-truth and prediction. The lower the SqRel score, the better. 

%\begin{equation}
%    \frac{1}{|T|} \sum_{y \in T} \frac{||y -\hat{y}||^2}{\hat{y}}
%\end{equation}

\textbf{Root Mean Square Error (RMS):} measures how far off each pixel is from   ground-truth values.  
A low RMS score indicates a better prediction.

%\begin{equation}
%    \sqrt{\frac{1}{|T|} \sum_{y \in T} ||y_i -\hat{y_i}||^2}
%\end{equation}

\textbf{Root Mean Square Error Log scale (RMSlog$_{10}$):} is similar to RMSE but in log scale. The lower, the better.
%\begin{equation}
%    \sqrt{\frac{1}{|T|} \sum_{y \in T} ||log y_i -log \hat{y_i}||^2}
%\end{equation}

\textbf{Threshold Accuracy ($\delta_r$):} indicates the accuracy of the prediction by setting a threshold for different ratios (r) between ground-truth and prediction. The higher, the better.
%\begin{equation}
%    max(\frac{y_i}{\hat{y}_i},\frac{\hat{y}_i}{y_i}) = \delta < thr
%\end{equation}

For the depth evaluation, we follow the previous methods~\cite{2019arXiv190810553B,2019arXiv190712209Y,2017arXiv170407813Z} and use the same evaluation protocols.

%%%%%%%%%%%%%%%%%%%%%%%%%%%%%%%%%%%%%%%%% TABLE Segmentation Results %%%%%%%%%%%%%%%%%%%%%%%%%%%%%%%%%%%%%%%
\begin{table*}[!ht]
\renewcommand*{\arraystretch}{1.12}
\centering
\resizebox{0.9\hsize}{!}{
\begin{tabular}{c||cccccccccccccc|c}
Approach &  \rotatebox{90}{Car} & \rotatebox{90}{Bicycle} & \rotatebox{90}{Motorcycle} & \rotatebox{90}{Truck}    & \rotatebox{90}{Other-Vehicle} & \rotatebox{90}{Person} & \rotatebox{90}{Road} & \rotatebox{90}{Sidewalk} & \rotatebox{90}{Building} & \rotatebox{90}{Fence} & \rotatebox{90}{Vegetation} & \rotatebox{90}{Terrain}  & \rotatebox{90}{Pole} & \rotatebox{90}{Traffic-Sign} & mIoU $\uparrow$ \\ 
\hline
Pix2Pix~\cite{pix2pix2017}           & 8.8                                & 0                                 & 0                                 & 0                                  & 0                               & 0                                 & 57.7                               & 15.7                               & 32.8                               & 12.5                               & 32.7                               & 14.8                               & 0.5                             & 0                                  & 12.5                                  \\
%TITAN-Net (without range data) & 31.6                                & 1.2                                & 0.7                                & 0.25                               & \textbf{0.1}                             & 0.8                                & 48.3                                & 17.0                                & 50.9                                & 19.4                                & 44.9                                & 27.8                                & 0.8                              & 0.4                                 & 17.4              \\

GauGAN~\cite{Park_2019_CVPR}                                                    & \multicolumn{1}{c}{42.9}               & \multicolumn{1}{c}{0}                   & \multicolumn{1}{c}{0}                      & \multicolumn{1}{c}{6.4}                 & \multicolumn{1}{c}{0}                         & \multicolumn{1}{c}{0}                  & \multicolumn{1}{c}{55.5}                & \multicolumn{1}{c}{31.3}                    & \multicolumn{1}{c}{36.8}                    & \multicolumn{1}{c}{11.7}                 & \multicolumn{1}{c}{38.7}                      & \multicolumn{1}{c}{18.7}                   & \multicolumn{1}{c}{0.07}                & \multicolumn{1}{c}{0.01}                                             & \multicolumn{1}{|c}{17.3}       \\
TITAN-Net~\cite{Cortinhal_2021_ICCV}                                          & \multicolumn{1}{c}{68.2}               & \multicolumn{1}{c}{9.9}                    & \multicolumn{1}{c}{7.6}                       & \multicolumn{1}{c}{7.6}                  & \multicolumn{1}{c}{0}                            & \multicolumn{1}{c}{7.3}                   & \multicolumn{1}{c}{75.4}                & \multicolumn{1}{c}{48.3}                    & \multicolumn{1}{c}{62.9}                    & \multicolumn{1}{c}{33.6}                 & \multicolumn{1}{c}{60.1}                      & \multicolumn{1}{c}{49.9}                   & \multicolumn{1}{c}{2}                   & \multicolumn{1}{c|}{3.7}                                             & 31.1 \\
%TITAN-Next (ours)$-$ w/o Depth                                                    & \multicolumn{1}{c}{85.9}               & \multicolumn{1}{c}{27.7}                   & \multicolumn{1}{c}{15.5}                      & \multicolumn{1}{c}{\textbf{18.8}}                 & \multicolumn{1}{c}{0.0}                         & \multicolumn{1}{c}{\textbf{43.1}}                  & \multicolumn{1}{c}{86.2}                & \multicolumn{1}{c}{63.5}                    & \multicolumn{1}{c}{76.2}                    & \multicolumn{1}{c}{51.0}                 & \multicolumn{1}{c}{73.4}                      & \multicolumn{1}{c}{66.2}                   & \multicolumn{1}{c}{20.9}                & \multicolumn{1}{c}{18.3}                                             & \multicolumn{1}{|c}{46.1} 

%\\

%TITAN-Next (ours)$-$ w/o SegMap                                                    & \multicolumn{1}{c}{83.4}               & \multicolumn{1}{c}{31.5}                   & \multicolumn{1}{c}{10.6}                      & \multicolumn{1}{c}{7.5}                 & \multicolumn{1}{c}{0.0}                         & \multicolumn{1}{c}{20.8}                  & \multicolumn{1}{c}{83.5}                & \multicolumn{1}{c}{54.7}                    & \multicolumn{1}{c}{68.7}                    & \multicolumn{1}{c}{41.7}                 & \multicolumn{1}{c}{67.6}                      & \multicolumn{1}{c}{62.0}                   & \multicolumn{1}{c}{14.6}                & \multicolumn{1}{c}{16.0}                                             & \multicolumn{1}{|c}{40.1} 
TITAN-Next w/o Depth & 85.0 & 38.7 & 18.3 & 16.6 & 0 & 29.3 & 89.3 & 67.2 & 70.0 & 52.6 & \textbf{74.3} & 66.9 & \textbf{21.4} & 21.7 & 46.5 \\
TITAN-Next (Ours)                                                    & \multicolumn{1}{c}{\textbf{86.6}}               & \multicolumn{1}{c}{\textbf{49.1}}                   & \multicolumn{1}{c}{\textbf{45.9}}                      & \multicolumn{1}{c}{\textbf{35.7}}                 & \multicolumn{1}{c}{\textbf{33.1}}                         & \multicolumn{1}{c}{\textbf{38.7}}                  & \multicolumn{1}{c}{\textbf{89.9}}                & \multicolumn{1}{c}{\textbf{68.5}}                    & \multicolumn{1}{c}{\textbf{71.9}}                    & \multicolumn{1}{c}{\textbf{53.2}}                 & \multicolumn{1}{c}{74.2}                      & \multicolumn{1}{c}{\textbf{70.0}}                   & \multicolumn{1}{c}{20.5}                & \multicolumn{1}{c}{\textbf{30.2}}                                             & \multicolumn{1}{|c}{\textbf{54.8}} 
\end{tabular}
}
\caption{Quantitative results for the generated camera   segments on the test sequences.  
 $\uparrow$ denotes that   higher is better.  }
   \label{tab:gen_sem_seg}
\end{table*}
%%%%%%%%%%%%%%%%%%%%%%%%%%%%%%%%%%%%%%%%%%%%%%%%%%%%%%%%%%%%%%%%%%%%%%%%%%%%%%%%

%%%%%%%%%%%%%%%%%%%%%%%%%%%%%%%%%%%%%%%%% TABLE RGB Results %%%%%%%%%%%%%%%%%%%%%%%%%%%%%%%%%%%%%%%

\begin{table*}[!ht]
\renewcommand*{\arraystretch}{1.12}
\centering
\resizebox{\textwidth}{!}{
\begin{tabular}{l||l|l|l|l|l|l|l|l|l}
\multicolumn{2}{l}{} & \multicolumn{8}{c}{SWD $\times 10^3$ $\downarrow$} \\ 
\cline{3-10}
Approach & FID $\downarrow$ & 1024$\times$1024 & 512$\times$512 & 256$\times$256 & 128$\times$128 & 64$\times$64 & 32$\times$32 & 16$\times$16 & avg \\ 
\hline
SC-UNET~\cite{Kim2020ColorIG}  & 261.28 & 2.65 & 2.56 & 2.36 & 2.20 & 2.14 & 2.14 & 4.07 & 2.59 \\
Pix2Pix~\cite{pix2pix2017} $\rightarrow$ Vid2Vid  & 73.47 & 2.29 & 2.22 & 2.18 & 2.15 & 2.11 & 2.13 & 3.95 & 2.59 \\
GauGAN~\cite{Park_2019_CVPR} $\rightarrow$Vid2Vid &94.35  & 2.25 & 2.18 & 2.14 & 2.14 & 2.11 & 2.14 & 4.07 & 2.43  \\ 
TITAN-Net~\cite{Cortinhal_2021_ICCV} $\rightarrow$ Vid2Vid  & 61.91 & 2.22 & 2.18 & 2.15 & 2.11& 2.08 & 2.13 & 3.78 & 2.38 \\

TITAN-Next (ours) $\rightarrow$Vid2Vid &\textbf{29.56}  & \textbf{2.18} & \textbf{1.88} & \textbf{1.90} & \textbf{1.86} & \textbf{1.76} & \textbf{1.57} & \textbf{1.46} & \textbf{1.82}  
\\[0.2em] 

\hline  
Pix2Pix~\cite{pix2pix2017}$-$ w/o SegMap & 209.15 & 2.43 & 2.35 & 2.31 & 2.31 & 2.35 & 2.42 & 4.76 & 2.71 \\
GauGAN~\cite{Park_2019_CVPR}$-$ w/o SegMap & 251.1 & 2.24 & 2.15 & 2.10 & 2.10 & 2.01 & 1.99 & 3.50 & 2.08 \\
TITAN-Net~\cite{Cortinhal_2021_ICCV} $-$ w/o SegMap & 326.29 & 3.24 & 3.14 & 2.98 & 2.61 & 2.23 & 2.10 & 3.94 & 2.89 \\
 TITAN-Next (ours)$-$ w/o SegMap & 300.3 & 1.99 & 1.94 & 1.89 & 1.84 & 1.79 & 1.79 & 3.22 & 2.27 \\
TITAN-Next (ours)$-$ w/o Depth & 312.13 & 4.38 & 4.26 & 4.16 & 4.04 & 3.93 & 3.93 & 7.03 & 4.53 \\
%\red{TITAN-Next (ours)$-$w/o Rangeview $\rightarrow$Vid2Vid} & 0.0 & 0.0 & 0.0 & 0.0 & 0.0 & 0.0 & 0.0 & 0.0 & 0.0 \\

\hline\hline%\\[-1.0em] 
\multicolumn{1}{l}{SD-Net~\cite{2020arXiv200510821T} $\rightarrow$  Vid2Vid} & \multicolumn{1}{l}{20.3694} & \multicolumn{1}{l}{2.10} & \multicolumn{1}{l}{1.99} & \multicolumn{1}{l}{1.86} & \multicolumn{1}{l}{1.71} & \multicolumn{1}{l}{1.56} & \multicolumn{1}{l}{1.41} & \multicolumn{1}{l}{1.70} & 1.76
\end{tabular}
 
}
\caption{Quantitative results for the synthesized RGB images using the test sequences. 
Each level of the Laplacian Pyramid corresponds to a given resolution. The distances are shown per level, and the average gives us the overall distance between both distributions. Due to the nature of SWD, both images are resized to $1024\times1024$ before calculating the distance.
 $\downarrow$ denotes that lower scores are better. 
 }
   \label{tab:gen_rgb_im}
\end{table*}	
%%%%%%%%%%%%%%%%%%%%%%%%%%%%%%%%%%%%%%%%%%%%%%%%%%%%%%%%%%%%%%%%%%%%%%%%%%%%%%%%

%%%%%%%%%%%%%%%%%%%%%%%%%%%%%%%%%%%%%%%%%%%%%%%%%%%%%%%%%%%%%%%%%%%%%%%%%%%%%%%%
\subsection{Baselines}
In addition to the original TITAN-Net~\cite{Cortinhal_2021_ICCV}, various baselines are used by following the same training protocols on the same dataset for a  fair performance evaluation.
When it comes to Image-to-Image translation, to the best of our knowledge, only~\cite{pix2pix2017,NIPS2016_502e4a16,NEURIPS2019_b2eb7349,2020arXiv200404977N,Park_2019_CVPR,CycleGAN2017} perform the mapping from  segmentation maps to RGB Images. 
We restrict our baselines to Pix2Pix~\cite{pix2pix2017} and GauGAN~\cite{Park_2019_CVPR} since their source codes are available. 

\textbf{Pix2Pix}~\cite{pix2pix2017} is a well-known generative model dedicated to image-to-image translation. To measure the quality of the generated segments and images, we replace TITAN-Next with Pix2Pix in the framework shown in~Fig.~\ref{fig:pipeline}.

\textbf{GauGAN}~\cite{Park_2019_CVPR} is another GAN model for image-to-image translation. We    follow the same strategy used for Pix2Pix. 

\textbf{SC-UNET}~\cite{Kim2020ColorIG} is a U-Net based generative model which is not relying on adversarial training. SC-UNET generates RGB scene images directly from raw point clouds without incorporating the semantic segments. 
Since the source code and data split are not released for public use, we here use our SC-UNET implementation to have a baseline for RGB image synthesis directly from LiDAR point clouds. 
Note that the here provided scores  substantially differ from these reported in~\cite{Kim2020ColorIG} since the original work in~\cite{Kim2020ColorIG} uses a very constricted subset of the dataset, which is not publicly available.

%%%%%%%%%%%%%%%%%%%%%%%%%%%%%%%%%%%%%%%%%%%%%%%%%%%%%%%%%%%%%%%%%%%%%%%%%%%%%%%%
\setlength{\columnsep}{0.1cm}
\begin{figure*}[!t]
\begin{multicols}{5}
    \begin{subfigure}{}
        \includegraphics[width=1.0\linewidth]{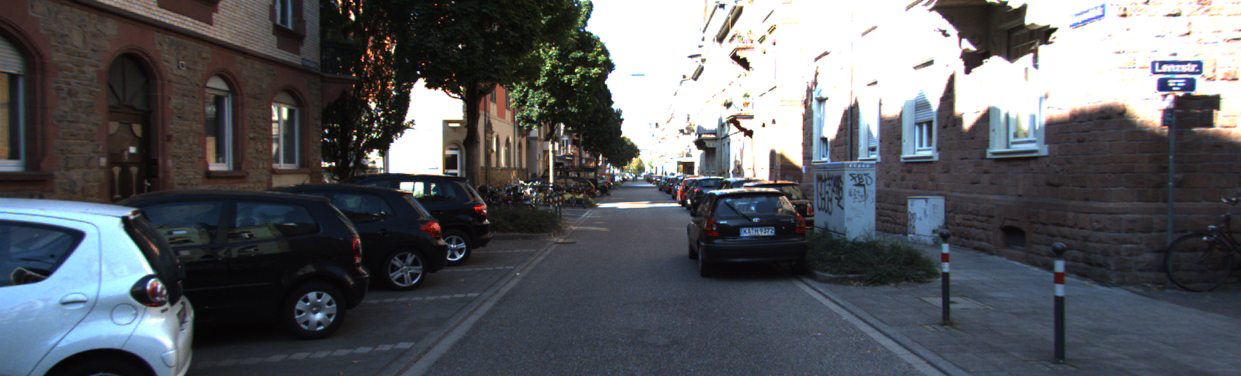} 
        \includegraphics[width=1.0\linewidth]{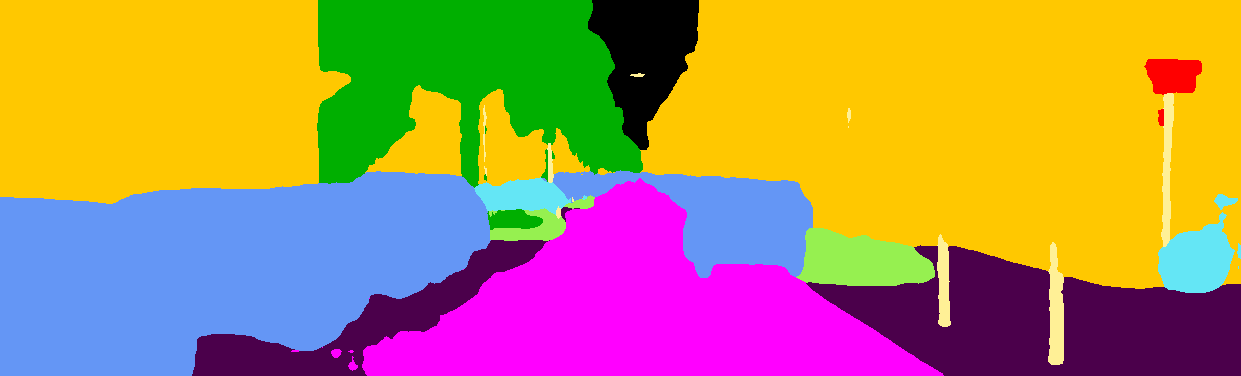}
        \includegraphics[width=1.0\linewidth]{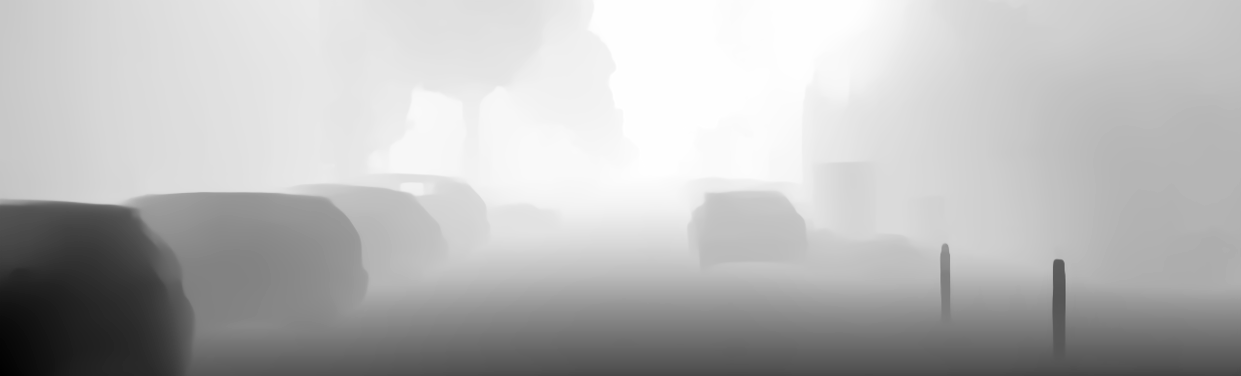}
        \includegraphics[width=1.0\linewidth]{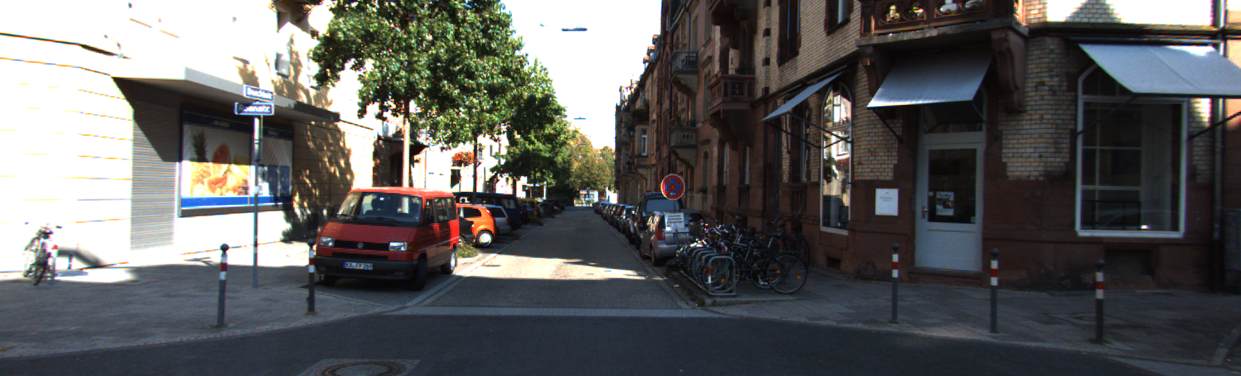}
        \includegraphics[width=1.0\linewidth]{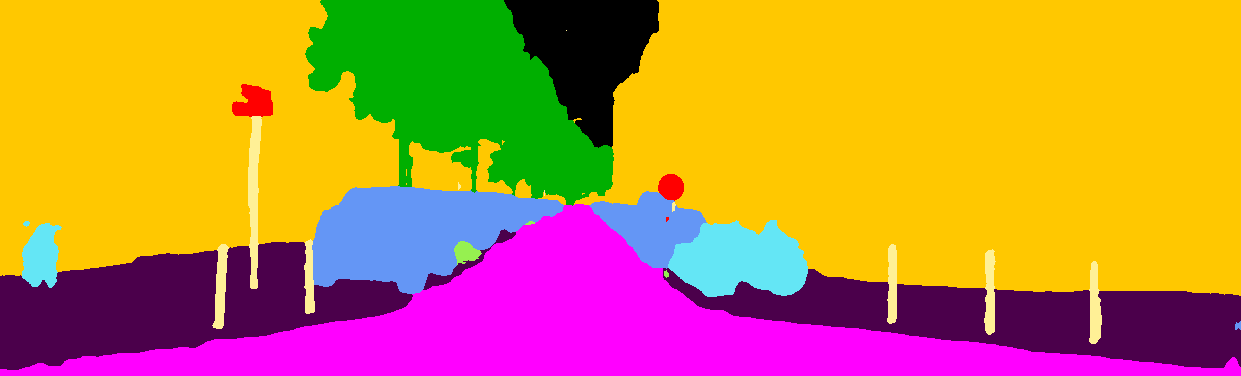}
        \includegraphics[width=1.0\linewidth]{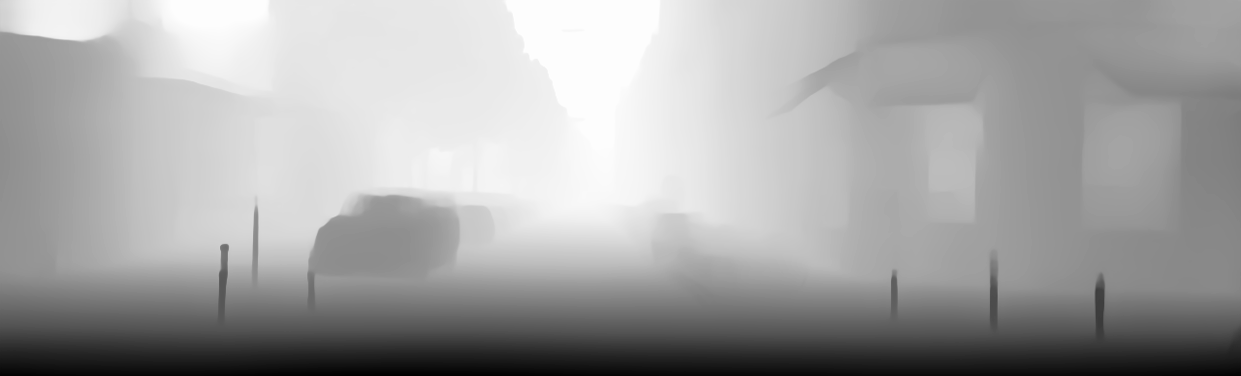}
        \caption*{Ground Truth}
    \end{subfigure}
    \begin{subfigure}{}
        \includegraphics[width=1.0\linewidth]{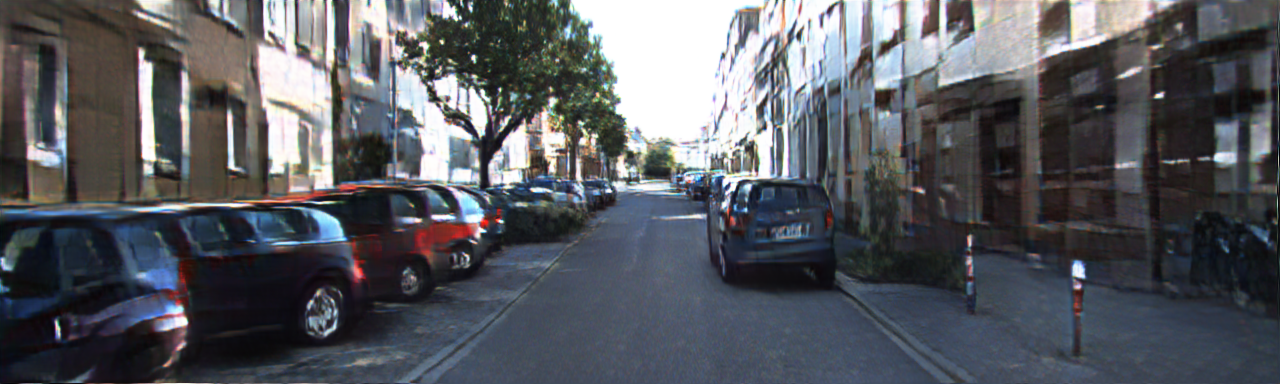}
        \includegraphics[width=1.0\linewidth]{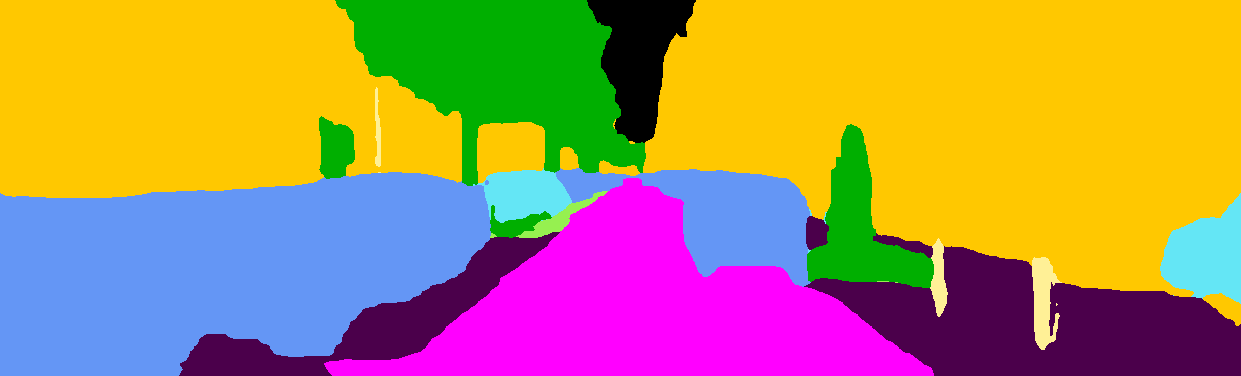}
        \includegraphics[width=1.0\linewidth]{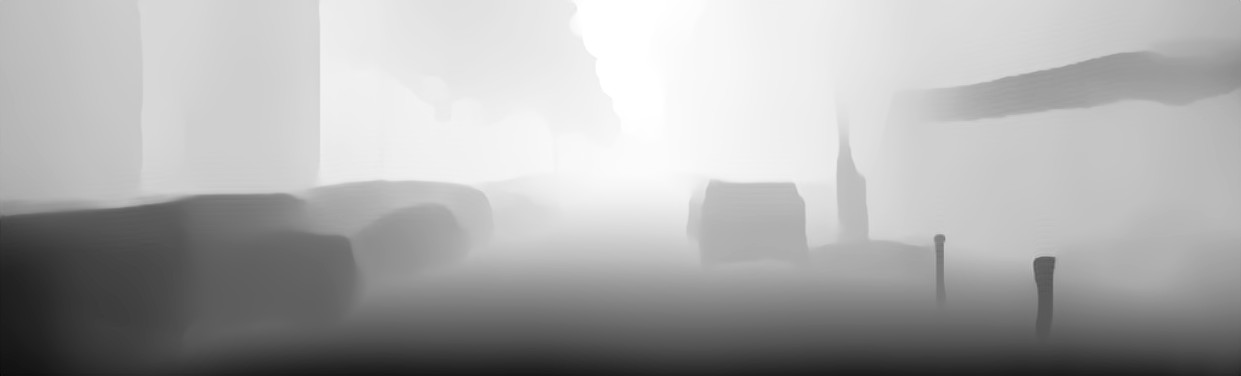}
        \includegraphics[width=1.0\linewidth]{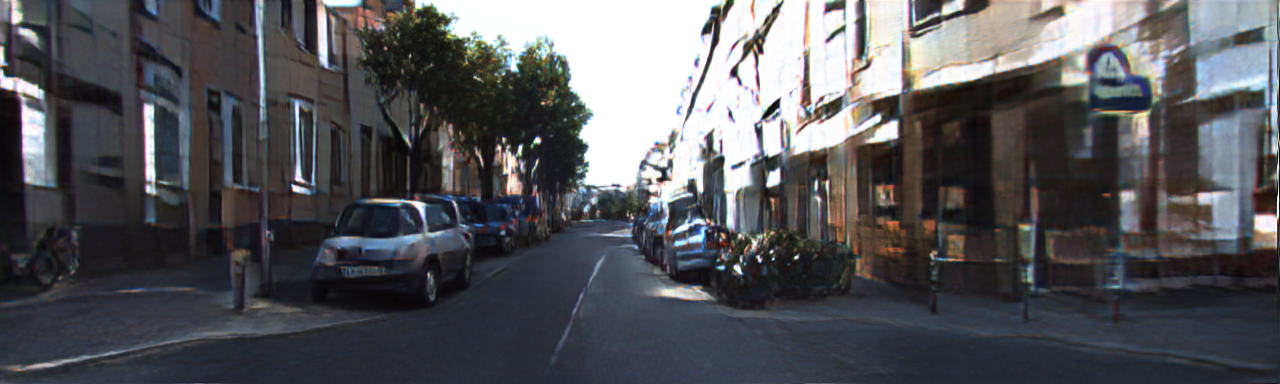}
        \includegraphics[width=1.0\linewidth]{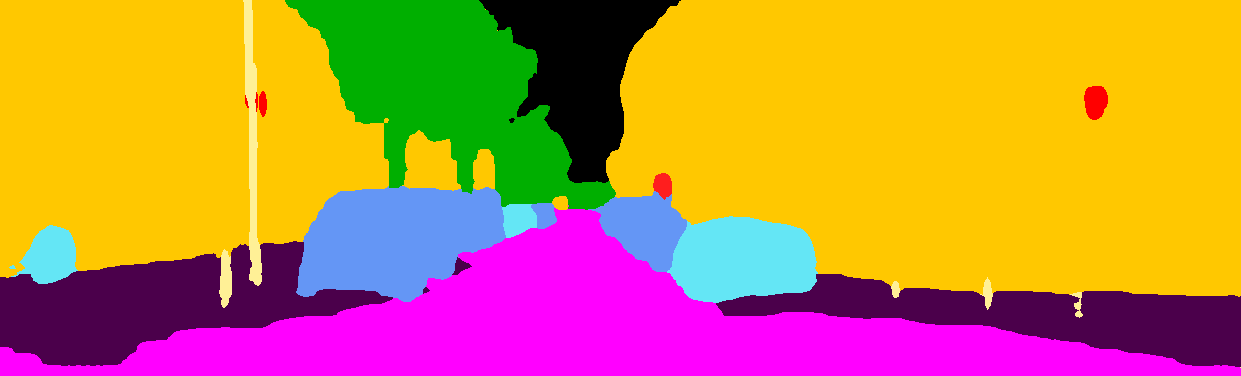}
        \includegraphics[width=1.0\linewidth]{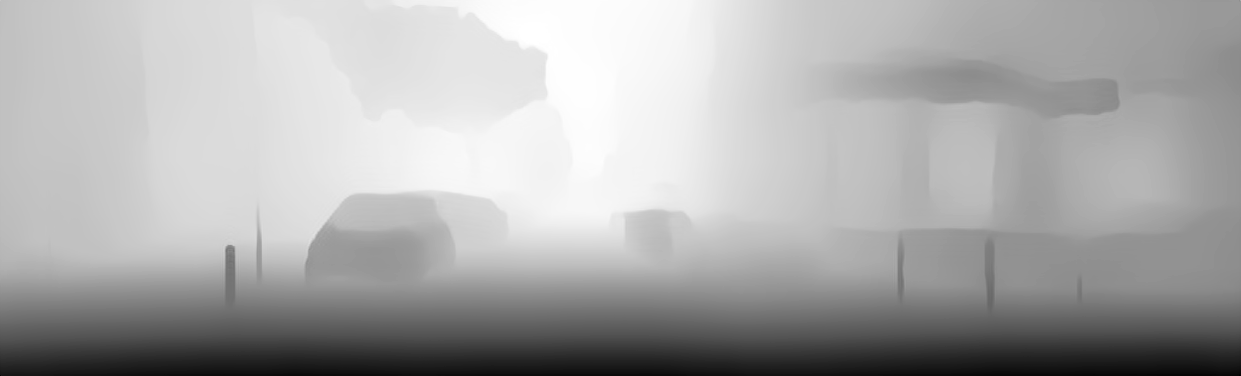}
        \caption*{TITAN-Next}
    \end{subfigure}
    
    \begin{subfigure}{}
        \includegraphics[width=1.0\linewidth]{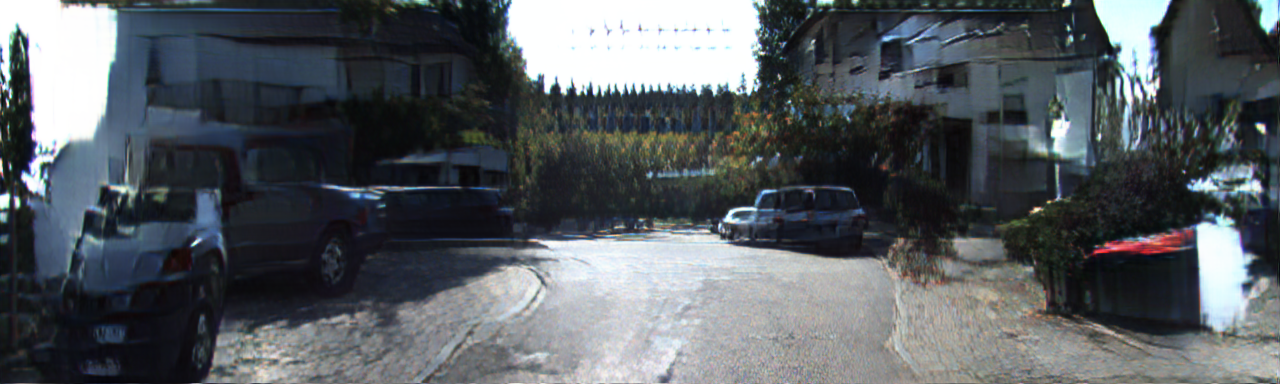}
        \includegraphics[width=1.0\linewidth]{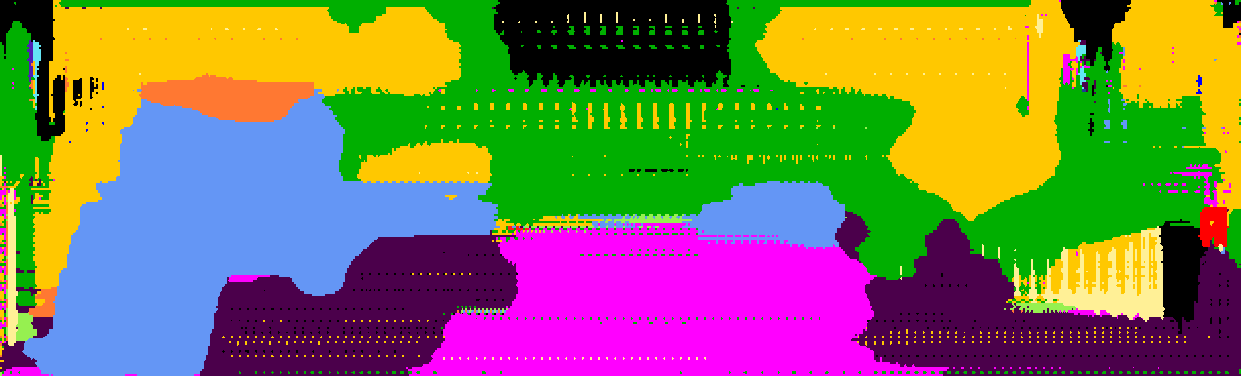}
        \includegraphics[width=1.0\linewidth]{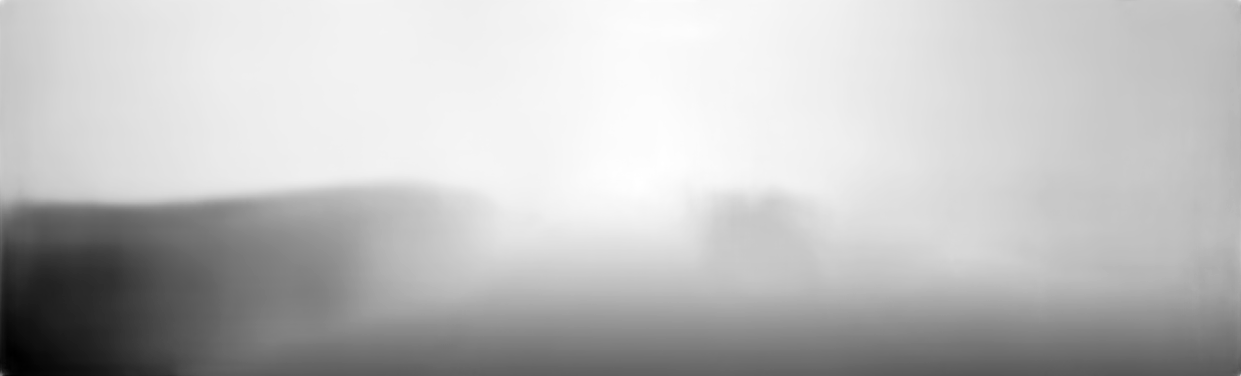}
        \includegraphics[width=1.0\linewidth]{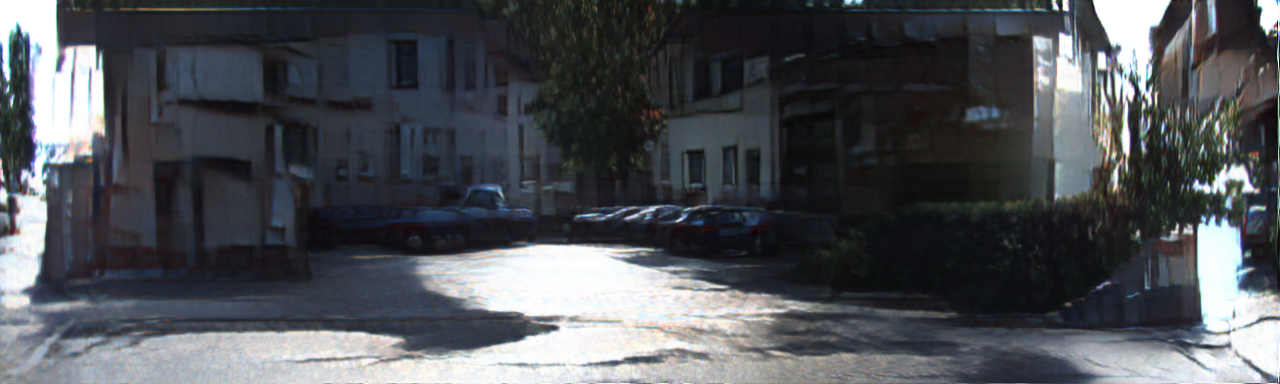}
        \includegraphics[width=1.0\linewidth]{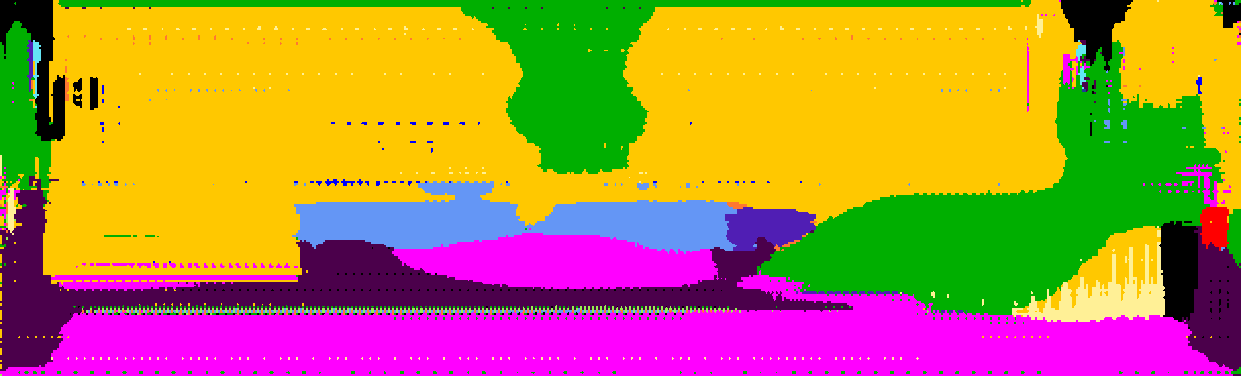}
        \includegraphics[width=1.0\linewidth]{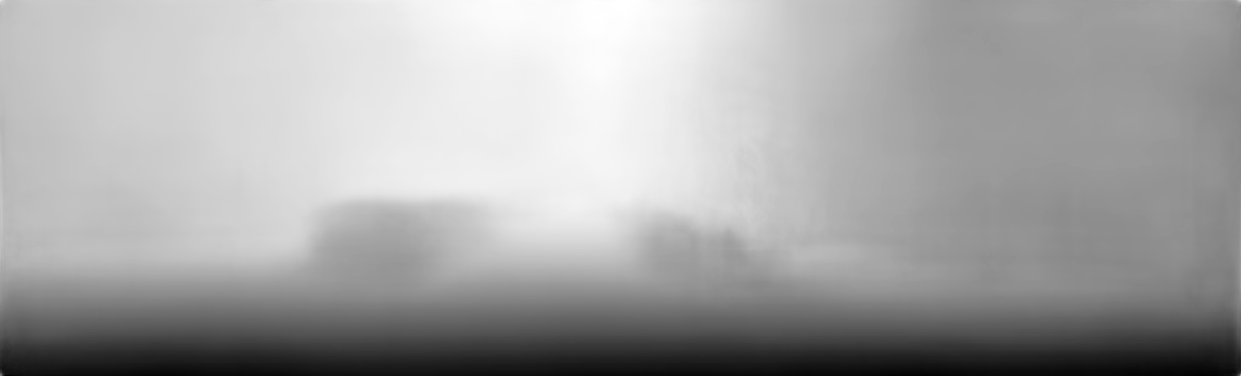}
        \caption*{GauGAN}
    \end{subfigure}
    
    \begin{subfigure}{}
        \includegraphics[width=1.0\linewidth]{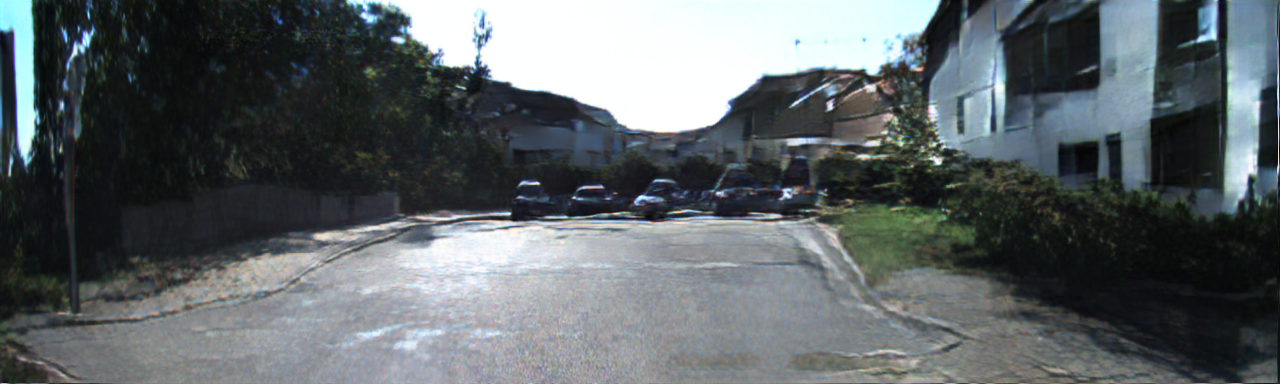}
        \includegraphics[width=1.0\linewidth]{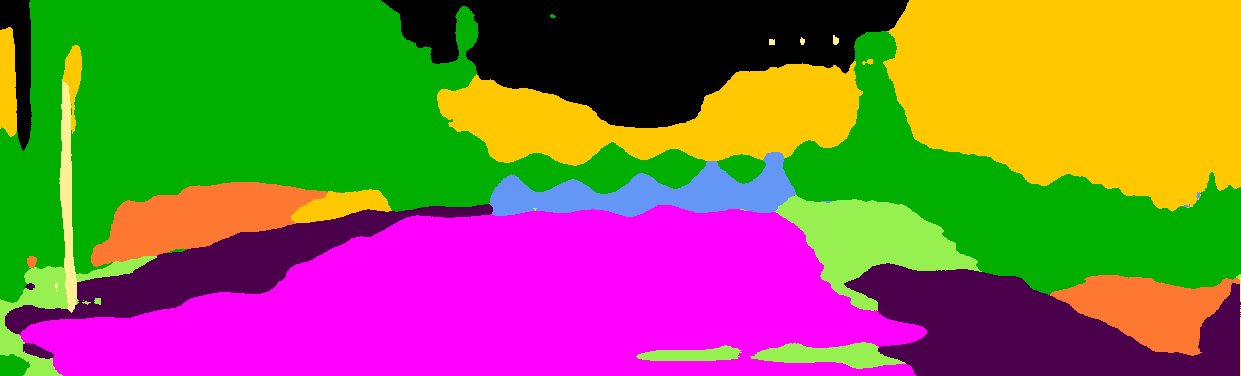}
        \includegraphics[width=1.0\linewidth]{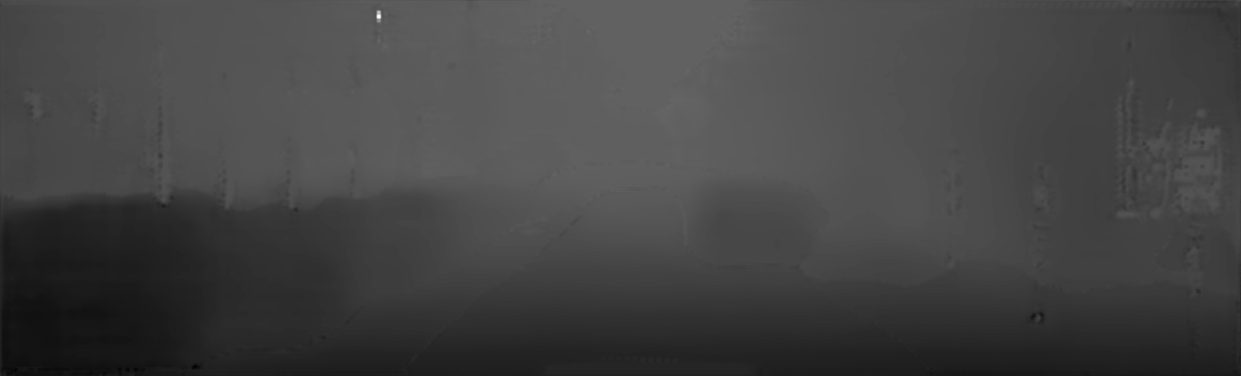}
        \includegraphics[width=1.0\linewidth]{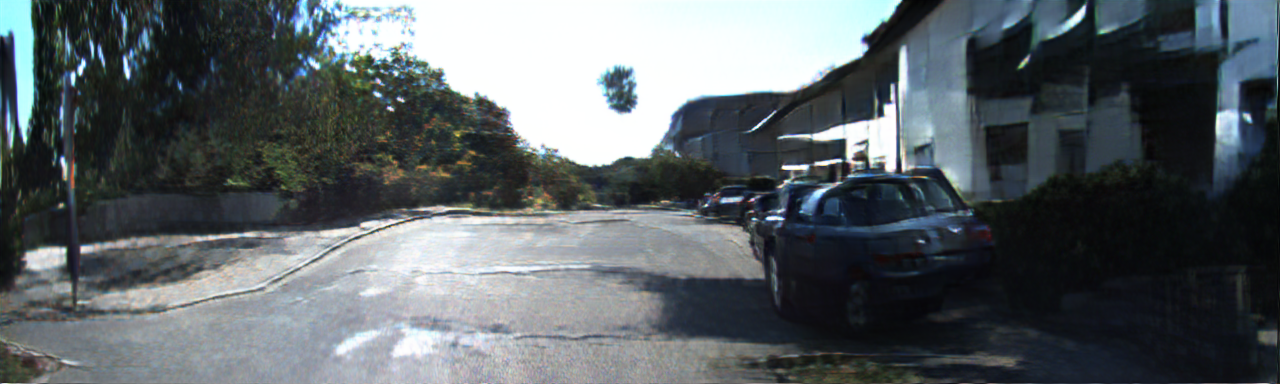}
        \includegraphics[width=1.0\linewidth]{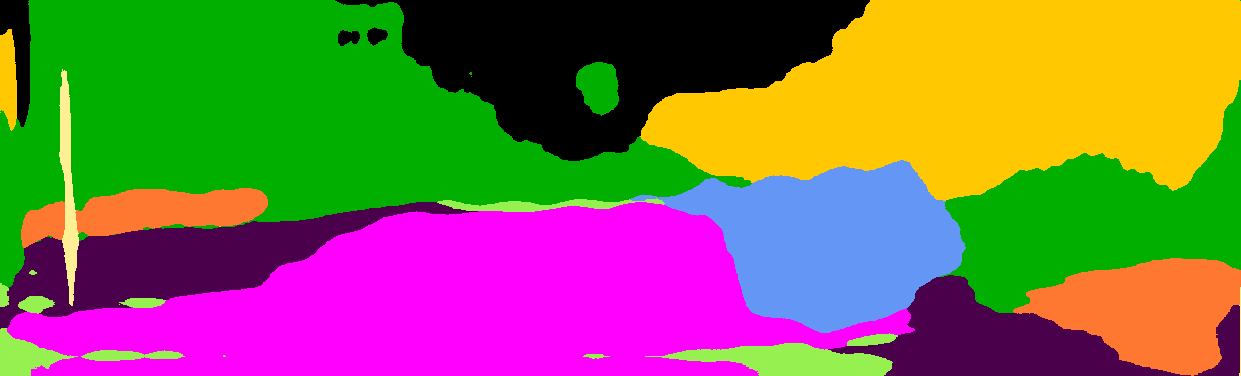}
        \includegraphics[width=1.0\linewidth]{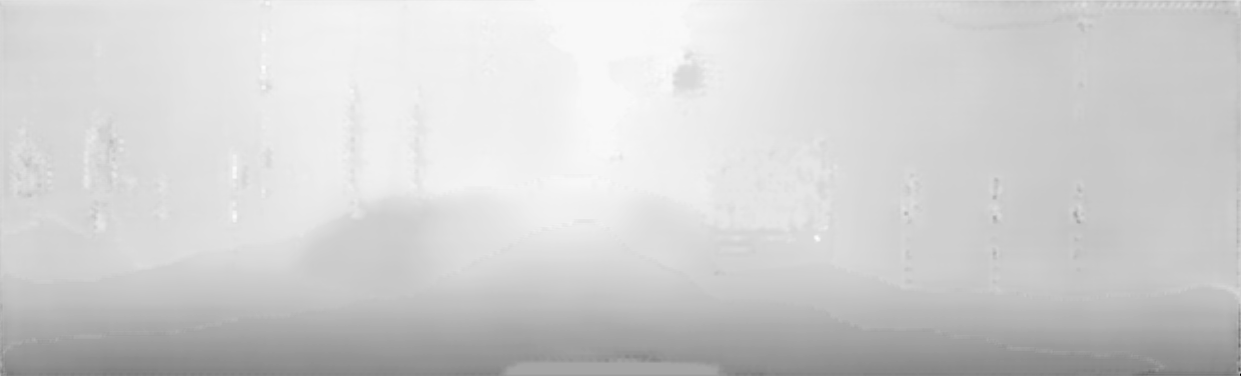}
        \caption*{Pix2Pix}
    \end{subfigure}

    \begin{subfigure}{}
        \includegraphics[width=1.0\linewidth]{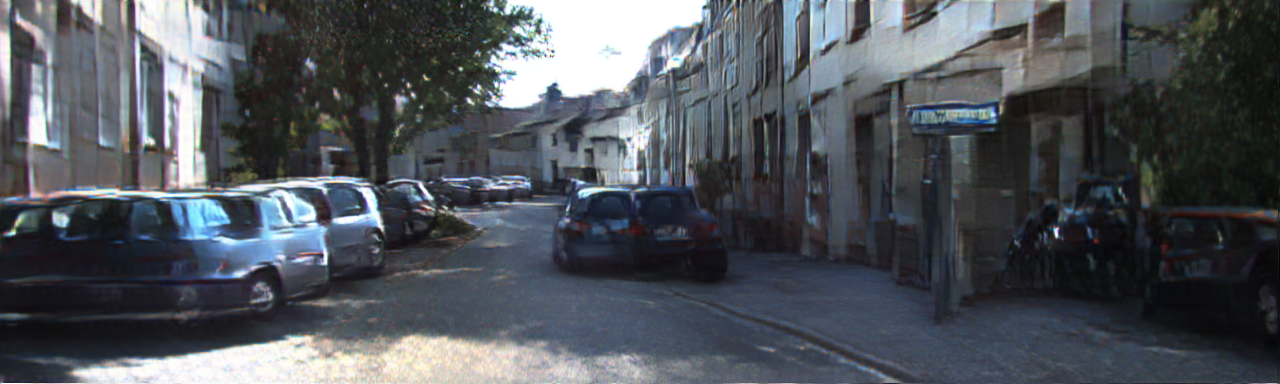}
        \includegraphics[width=1.0\linewidth]{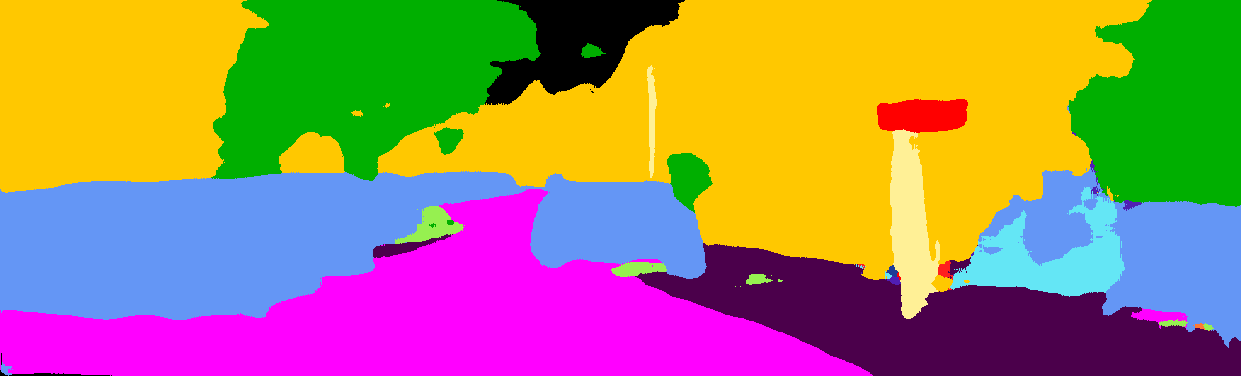}
        \includegraphics[width=1.0\linewidth]{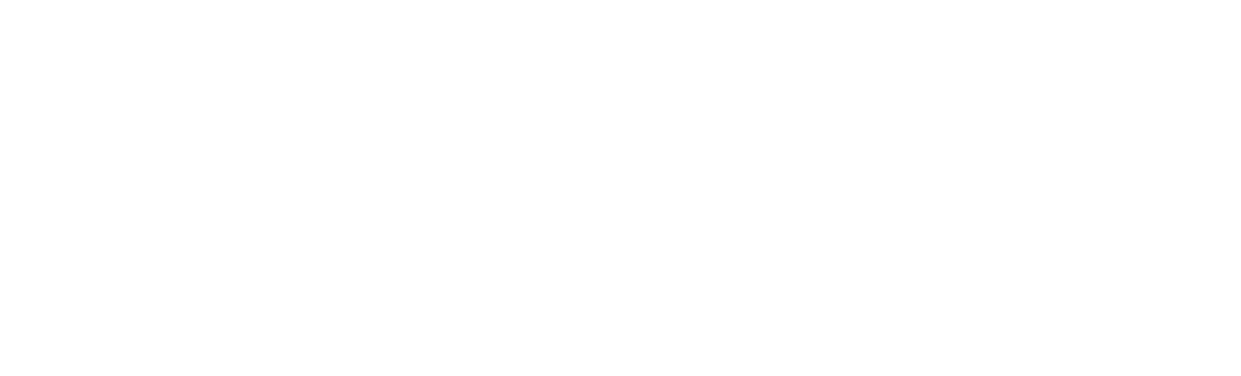}
        \includegraphics[width=1.0\linewidth]{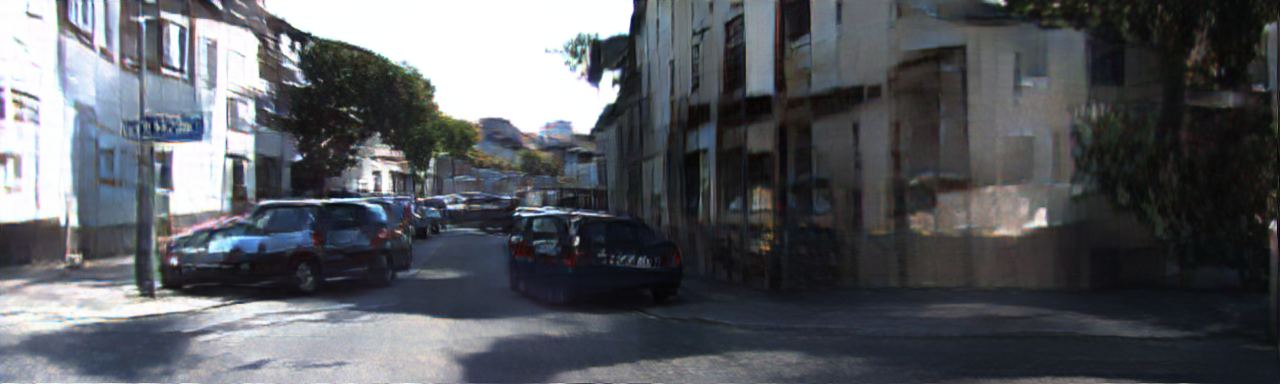}
        \includegraphics[width=1.0\linewidth]{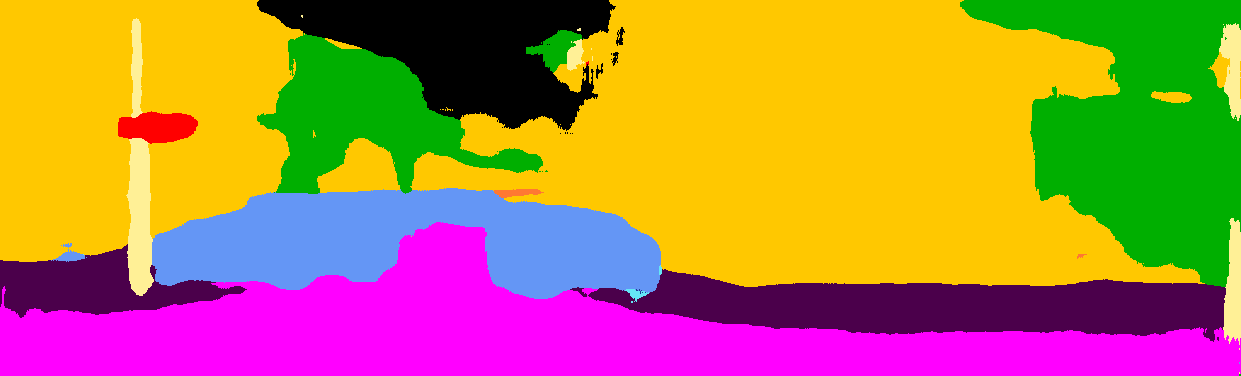}
        \includegraphics[width=1.0\linewidth]{placeholder.png}
        \caption*{TITAN-Net}
    \end{subfigure}

\end{multicols}
%\begin{multicols}{3}
%    \includegraphics[width=1.14\linewidth]{./images/qualitative/validation/real_rgb/000245.png}\par 
%    \includegraphics[width=1.14\linewidth]{./images/qualitative/validation/titan/rgb/000245.png}\par
%    \includegraphics[width=1.14\linewidth]{./images/qualitative/validation/pix2pix/rgb/000245.png}\par
%\end{multicols}
%\vspace{-0.8cm}
%\begin{multicols}{3}
%   \includegraphics[width=1.14\linewidth]{./images/qualitative/validation/real_segmaps/000245.png}\par 
%    \includegraphics[width=1.14\linewidth]{./images/qualitative/validation/titan/segmaps/000245.png}\par
%    \includegraphics[width=1.14\linewidth]{./images/qualitative/validation/pix2pix/segmaps/000245.png}\par
%\end{multicols}

\caption{Sample  synthesized images (at the top) with  the corresponding generated   segment masks and depth maps. From left to right, we have the ground-truth images and predictions from \textbf{TITAN-Next}, \textbf{GauGAN}~\cite{Park_2019_CVPR}, \textbf{Pix2Pix}~\cite{pix2pix2017} and \textbf{TITAN-Net}~\cite{Cortinhal_2021_ICCV}. Note that all these models are combined with Vid2Vid~\cite{wang2018vid2vid} to translate segments to RGB images.
}
\label{fig:synt_img_results}
\end{figure*}

\section{Quantitative and Qualitative Results}
  
The RGB-D image synthesis in our framework solely relies on the generated semantic scene segments. 
Therefore, we first start with quantitatively evaluating the segment maps $\widehat{y}_{s}$ generated from   $p^{\prime}$ and $p^{\prime}_{s}$ as shown in~Fig.~\ref{fig:pipeline}.

Table~\ref{tab:gen_sem_seg}  reports the mean IoU scores for the generated camera segmentation maps  on the \sk test sequences. 
We here note  that the ground truth segments are those generated by the SD-Net~\cite{2020arXiv200510821T}, i.e., $y_s$, which are also used by all the baseline and TITAN-Next discriminators. 
Table~\ref{tab:gen_sem_seg} clearly shows the substantial performance boost that our model TITAN-Next leads to. In contrast to the original TITAN-Net base model, we obtain a massive accuracy enhancement, i.e., $23.7\%$  improvement in the mean IoU score. 
There is a certain improvement for each individual segment class. 
In particular, a significant boost is obtained for rare classes such as bicycle, motorcycle, person, pole, among others.
When it comes to other baseline models Pix2Pix~\cite{pix2pix2017} and GauGAN~\cite{Park_2019_CVPR}, the segmentation performance gap gets even larger, i.e., TITAN-Next  considerably outperforms the others.

Table~\ref{tab:gen_rgb_im} presents  quantitative evaluations for the synthesized RGB images, $\widehat{y}$, on the \sk test sequences. 
As shown in Fig.~\ref{fig:pipeline}, TITAN-Next is combined with Vid2Vid (i.e., TITAN-Next $\rightarrow$ Vid2Vid) to  render the RGB scene images. 
Likewise, all the other baseline models are coupled with Vid2Vid, e.g.,  GauGAN $\rightarrow$ Vid2Vid, for a fair comparison. This, however, does not apply to the baseline  SC-UNET~\cite{Kim2020ColorIG}, since this model can generate scene images directly from the raw point cloud. 
The results in Table~\ref{tab:gen_rgb_im} indicate that our TITAN-Next model (i.e., TITAN-Next $\rightarrow$ Vid2Vid) substantially reduces the
FID and SWD scores, thus, considerably outperforms the others. 
At the bottom of Table~\ref{tab:gen_rgb_im}, we also provide the obtained scores for SD-Net~\cite{2020arXiv200510821T} $\rightarrow$ Vid2Vid, since SD-Net~\cite{2020arXiv200510821T} is mainly employed to generate the ground truth semantic segments. This way, we can clearly measure how much the synthesized images by our model (TITAN-Next $\rightarrow$ Vid2Vid) deviate from the expected image quality. In contrast to  other baselines, our TITAN-Next model returns the closest scores to these reference scores.  

Regarding the evaluation of the generated depth cues, Table~\ref{tab:depth} reports the obtained quantitative depth estimation results on the \sk test set. As clearly shown in Table~\ref{tab:depth}, there is a substantial improvement in all metrics in contrast to the baseline models, combined with an additional depth branch for the sake of fairness.

Fig.~\ref{fig:synt_img_results} depicts sample   synthesized images together with the corresponding generated segmentation masks and depth maps.
This figure ensures that the reconstructed semantic segments and depth maps by our TITAN-Next model are more accurate and, thus, the synthesized RGB images are more realistic compared to other baselines. 
TITAN-Next has a better reconstruction capability,  particularly for rare classes such as sidewalk, pole, and traffic sign which are  generated with high fidelity. 
Note that for a fair comparison, in this figure, we exclude results for SC-UNET~\cite{Kim2020ColorIG} since it bypasses the segmentation step.

%for RGB depth estimation, we used the pseudo labels $y_d$ coming from MiDaS~\cite{Ranftl2020}. Table \ref{tab:depth} shows the results on the test set of \sk.

%%%%%%%%%%%%%%%%%%%%%%%%%%%%%%%%%%%%%%%%%%%%%%%%%%%%%%%%%%%%%%%%%%%%%%%%%%%%%%%%

%%%%%%%%%%%%%%%%%%%%%%%%%%%%%%%%%%%%%%%%%%%%%%%%%%%%%%%%%%%%%%%%%%%%%%%%%%%%%%%%
\begin{figure*}[!t]
\begin{multicols}{5}
    \begin{subfigure}{}
        \includegraphics[width=1.0\linewidth]{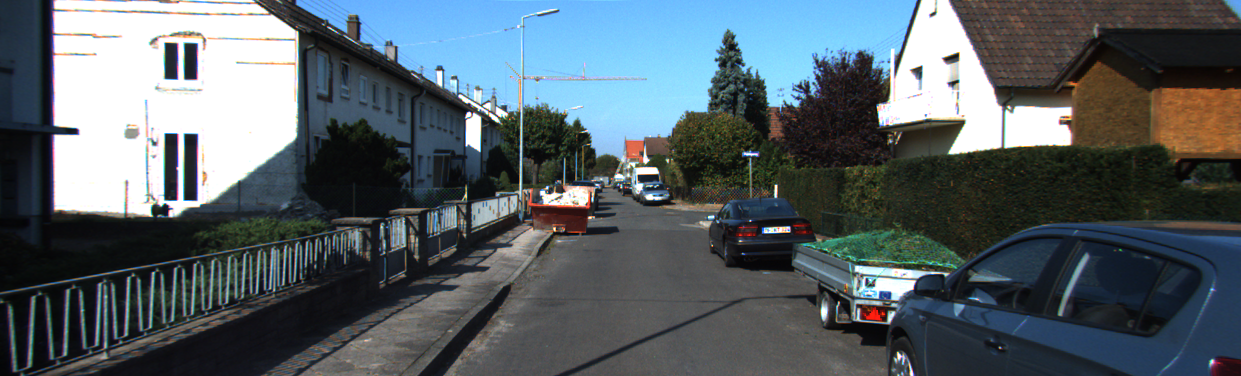} 
        \includegraphics[width=1.0\linewidth]{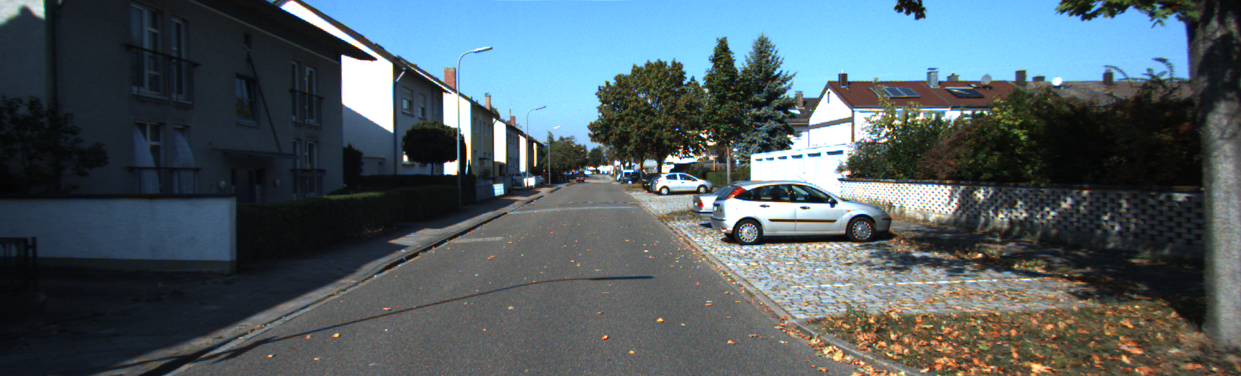}
        \includegraphics[width=1.0\linewidth]{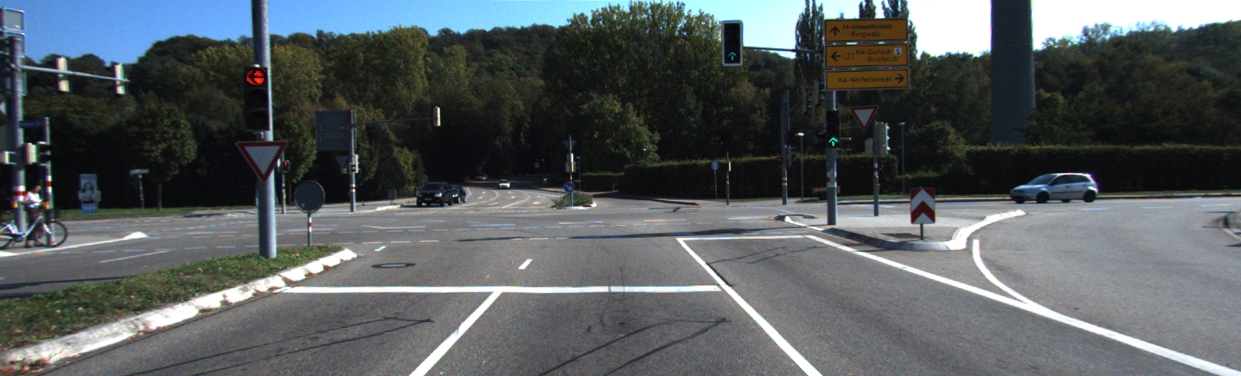} 
        \caption*{Ground Truth}
    \end{subfigure}
    %%%%%%%%%%%%%%%%%%%%%%%%%%%%%%%%%%%%%%%%%%%%%%%
     \begin{subfigure}{}
        \includegraphics[width=1.0\linewidth]{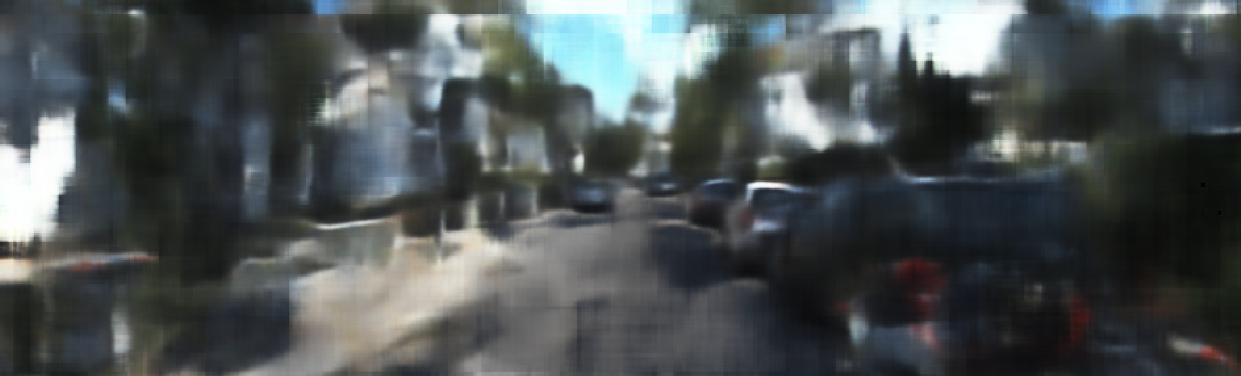}
        \includegraphics[width=1.0\linewidth]{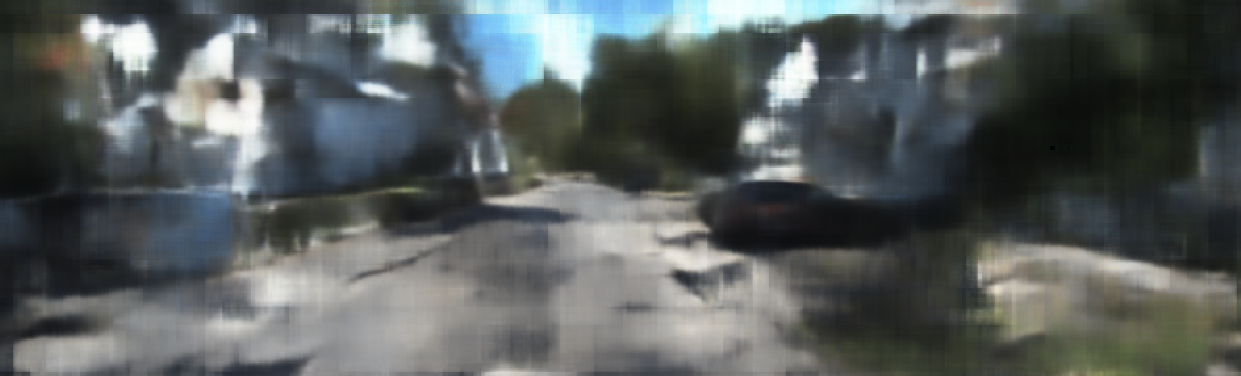}
        \includegraphics[width=1.0\linewidth]{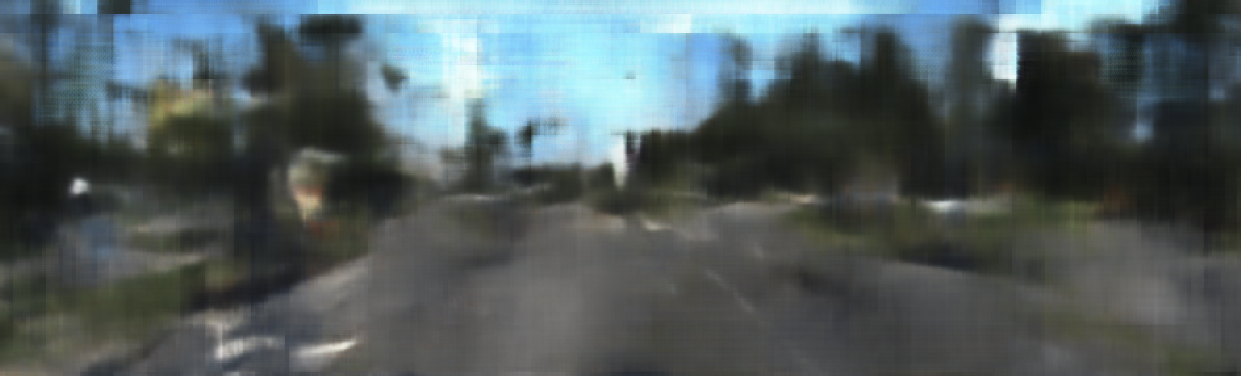}
        \caption*{SC-UNET}
    \end{subfigure}
    %%%%%%%%%%%%%%%%%%%%%%%%%%%%%%%%%%%%%%%%%%%%%%%
    \begin{subfigure}{}
        \includegraphics[width=1.0\linewidth]{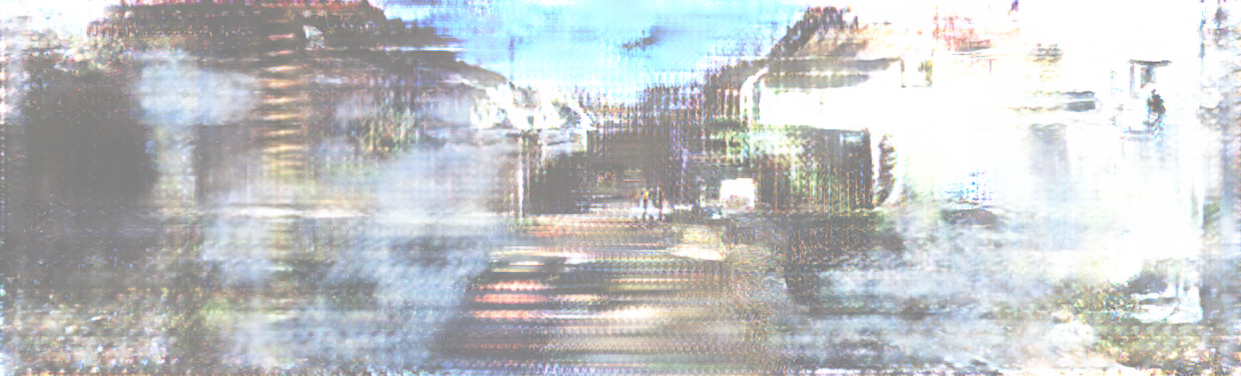}
        \includegraphics[width=1.0\linewidth]{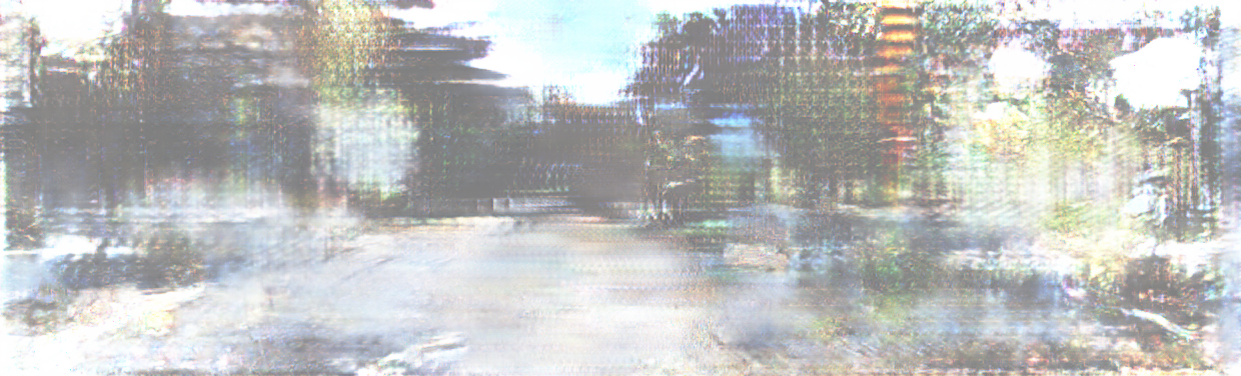}
        \includegraphics[width=1.0\linewidth]{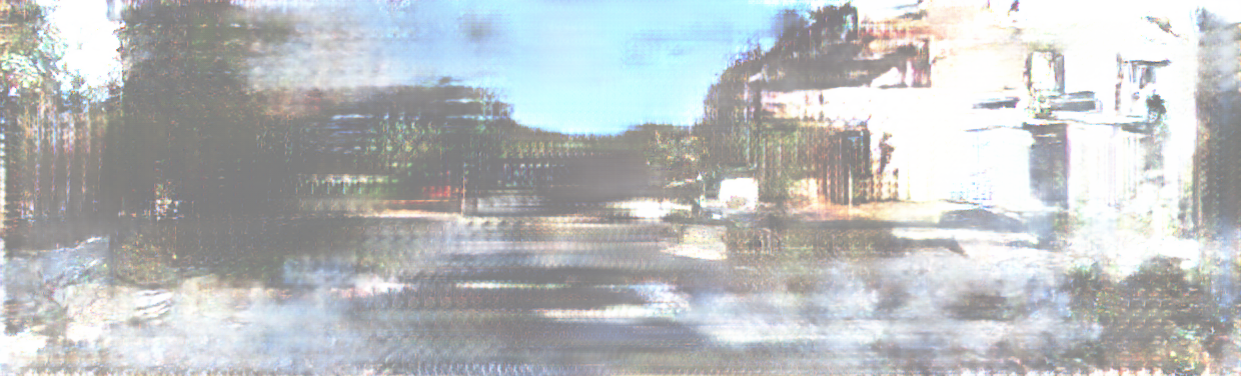}
        \caption*{Pix2Pix}
    \end{subfigure}
    %%%%%%%%%%%%%%%%%%%%%%%%%%%%%%%%%%%%%%%%%%%%%%%
    \begin{subfigure}{}
        \includegraphics[width=1.0\linewidth]{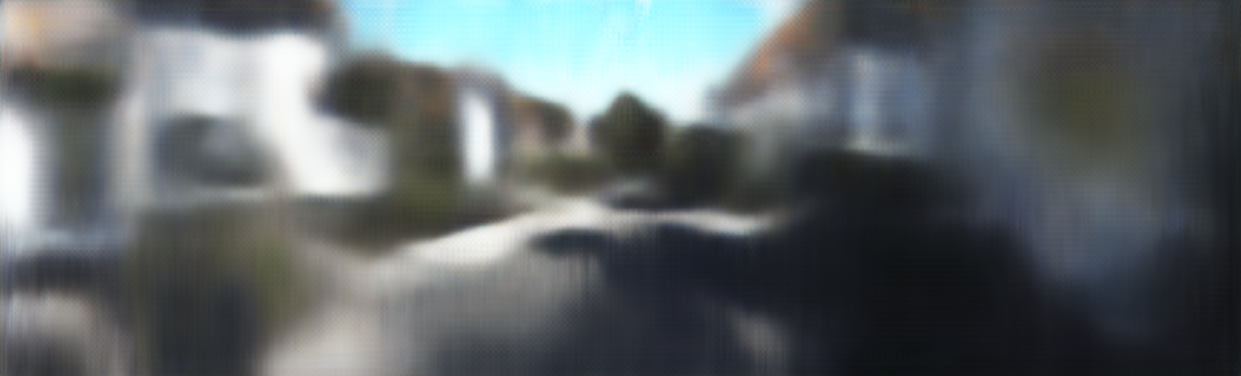}
        \includegraphics[width=1.0\linewidth]{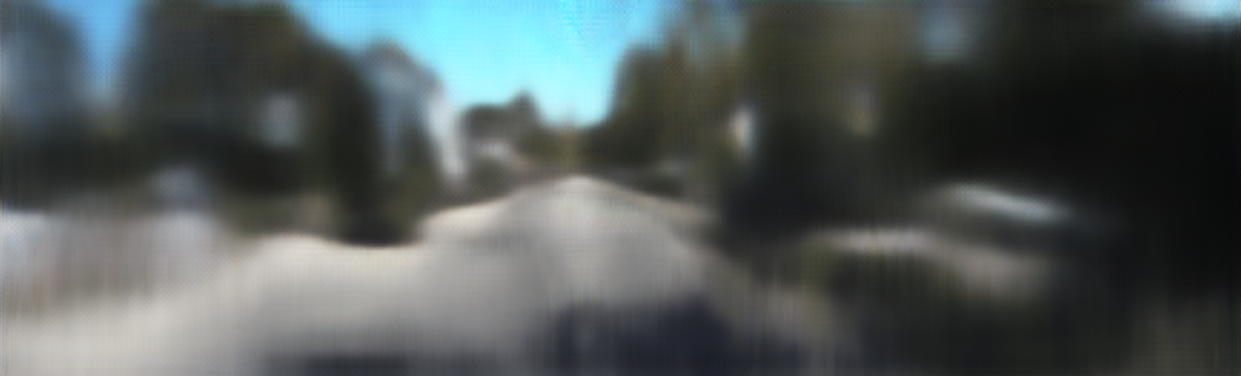}
        \includegraphics[width=1.0\linewidth]{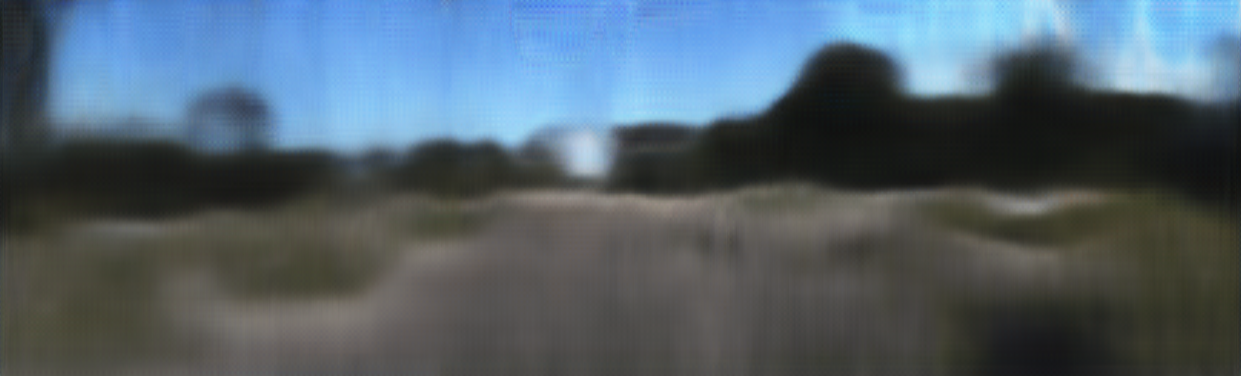}
        \caption*{TITAN-Net}
    \end{subfigure}
    %%%%%%%%%%%%%%%%%%%%%%%%%%%%%%%%%%%%%%%%%%%%%%%
    \begin{subfigure}{}
        \includegraphics[width=1.0\linewidth]{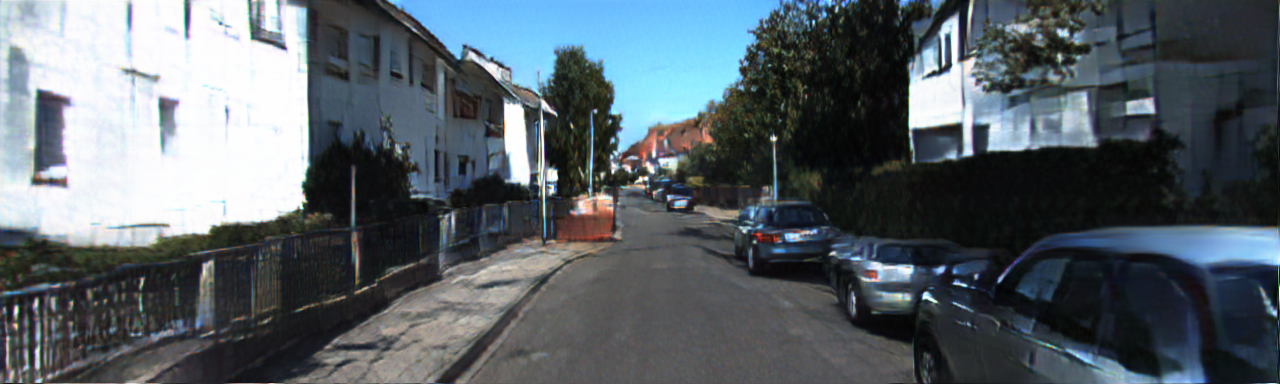}
        \includegraphics[width=1.0\linewidth]{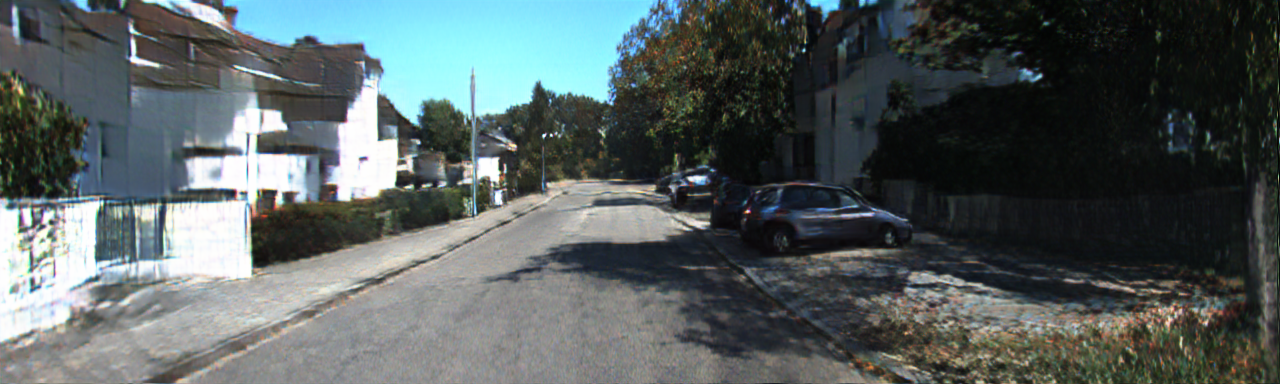}
        \includegraphics[width=1.0\linewidth]{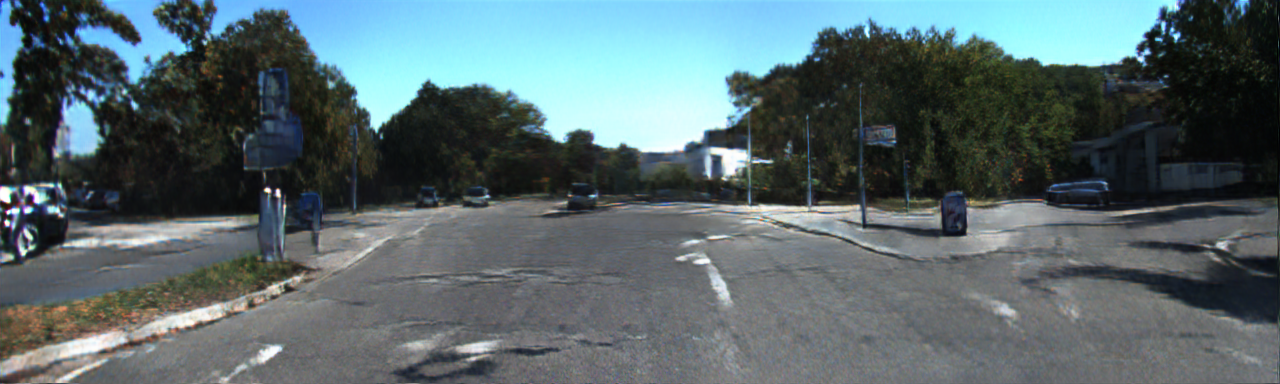}
        \caption*{TITAN-Next}
    \end{subfigure}

\end{multicols}
\caption{Sample synthesized images with the corresponding images trained on \textbf{SC-UNET}\cite{Kim2020ColorIG}, \textbf{Pix2Pix}\cite{pix2pix2017},    \textbf{TITAN-Net}\cite{Cortinhal_2021_ICCV} and \textbf{TITAN-Next (ours)}, without the semantic maps as an intermediary representation. %First images of each row are the ground-truth.
}
\label{fig:wosemantic}
\end{figure*}

%%%%%%%%%%%%%%%%%%%%%%%%%%%%%%%%%%%%%%%%%%%%%%%%%%%%%%%%%%%%%%%%%%%%%%%%%%%%%%%%

% \usepackage{booktabs}
\begin{table}[!b]
\centering
\resizebox{0.99\hsize}{!}{\begin{tabular}{@{}lcccc|ccc@{}}
\toprule
            & AbsRel $\downarrow$ & SqRel $\downarrow$ & RMS $\downarrow$  & RMSlog$_{10}$ $\downarrow$& $\delta_1$ $\uparrow$  & $\delta_2$ $\uparrow$  & $\delta_3$ $\uparrow$ \\  

Pix2Pix + Depth & 123.73  & 1119.50 & 14.24& 2.19 & 0.16        & 0.40      & 0.40      \\
GauGAN + Depth & 110.06 & 637.50 & 3.72 & 1.42 & 0.63 & 0.87 & 0.87  \\
\textbf{TITAN-Next (ours)} & \textbf{43.14}  & \textbf{323.22} & \textbf{3.08} & \textbf{1.27}         & \textbf{0.75}       & \textbf{0.92}       & \textbf{0.92} \\
\bottomrule
\end{tabular}}
\caption{Quantitative results for the depth estimation. $\downarrow$ denotes lower is better while $\uparrow$ denotes higher is better.}
\label{tab:depth}
\end{table}
%%%%%%%%%%%%%%%%%%%%%%%%%%%%%%%%%%%%%%%%%%%%%%%%%%%%%%%%%%%%%%%%%%%%%%%%%%%%%%%%
%%%%%%%%%%%%%%%%%%%%%%%%%%%%%%%%%%%%%%%%%%%%%%%%%%%%%%%%%%%%%%%%%%%%%%%%%%%%%%%%
\subsection{Ablation Study}
In this section, we ablate each of our contributions  and  investigate the individual improvements over the original TITAN-Net
model. 
Table \ref{tab:ablation} reports the mean IoU scores  on the \sk test set together with the total parameter numbers in the generator network.
This table clearly shows that each contribution to TITAN-Net leads to a certain improvement in the model performance.  
%

%%%%%%%%%%%%%%%%%%%%%%%%%%%%%%%%%%%%%%%%%%%%%%%%%%%%%%%%%%%%%%%%%%%%%%%%%%%%%%%%
\begin{table}[!t]
\resizebox{0.99\hsize}{!}{
\begin{tabular}{@{}lccc@{}}
\toprule
Network              & \begin{tabular}{@{}c@{}}Number of\\ Generator Parameters   \end{tabular}& mIoU$\uparrow$ & FLOPs \\ \midrule
TITAN-Net~\cite{Cortinhal_2021_ICCV}     &     6.73M     &   31.1 &  20.58G\\
+ Encoder Feature Pyramid                &     7.22M     &  46.0 &    64.58G\\
+ Decoder Feature Pyramid                &     7.22M     &   54.8 & 67.88G  \\ \bottomrule
\end{tabular}
}
\caption{Ablative analysis. $\uparrow$ means  higher the better. }
   \label{tab:ablation}
\end{table}
%%%%%%%%%%%%%%%%%%%%%%%%%%%%%%%%%%%%%%%%%%%%%%%%%%%%%%%%%%%%%%%%%%%%%%%%%%%%%%%%

Furthermore, we pose the following questions to better study the network performance:

\textbf{What if we exclude the semantic segmentation step?}  
This question is crucial to explore the role of semantic segments in the  RGB image synthesis.
To assess this role, we measure the performance of TITAN-Next and other baseline models once the RGB images ($\widehat{y}$) are directly generated from the raw LiDAR point cloud projections ($p^{\prime}$) without incorporating the semantic segments, as in the case of SC-UNET~\cite{Kim2020ColorIG}. 

Table~\ref{tab:gen_rgb_im} presents the obtained results, which clearly indicate that without the intermediate segmentation cues (\textit{w/o SegMap}) the image reconstruction performance for all models dramatically drops. 
Fig.\ref{fig:wosemantic} shows sample synthesized images where the quality is  drastically diminished.

%These results convey the fact that excluding the semantics drastically diminishes the quality of synthesized images.

%\textbf{What if TITAN-Next is not conditioned on  $p^{\prime}$?}  
%In the previous ablation study, we implicitly investigated the  role of LiDAR segments $p^{\prime}_{s}$ in image synthesis. We now   diagnose the contribution of $p^{\prime}$  in the quality of generated RGB images. The last row in Table~\ref{tab:gen_rgb_im} shows that when the condition on $p^{\prime}$ is removed, the obtained results get worse, in contrast to the results in the third row.

\textbf{What if we exclude the depth head?}
Depth cues can be an important factor in helping the network gain a more complete understanding of the scene. To examine the impact of depth cues, we trained TITAN-Next without the depth head (\textit{w/o Depth}) and evaluated the results, which are presented in Table~\ref{tab:gen_sem_seg} and Table~\ref{tab:gen_rgb_im}.

Our results show that including the depth head leads to a $8.3\%$ improvement in semantic segmentation performance (see Table~\ref{tab:gen_sem_seg}) and a substantial improvement in the synthesized images quality (see Table~\ref{tab:gen_rgb_im}).

%%%%%%%%%%%%%%%%%%%%%%%%%%%%%%%%%%%%%%%%%%%%%%%%%%%%%%%%%%%%%%%%%%%%%%%%%%%%%%%%
\subsection{Runtime}

The overall training time on two Quadros RTX 6000 GPUs is around five days for 144 epochs and 19K samples. 
On a single  Quadros RTX 6000  GPU, the inference runtime for point cloud segmentation, image segmentation together with depth estimation, translation between segmentation maps, and synthesizing a 376 $\times$  1241 image takes 40.89, 198.75, 41.20, and 123.72 msec, respectively.

\section{Limitations and Discussion}

At the core of this work, we hypothesize that the scene semantics can bridge the gap across different domains with distinct data structures, such as unorganized sparse 3D   point clouds and structured 2D image data.  
Our experimental findings reported in Table~\ref{tab:gen_rgb_im} and Fig.~\ref{fig:synt_img_results} reveal the fact that our hypothesis holds true that semantic scene segments, to a great extent, overcome the domain translation problem between the LiDAR and RGB camera modalities.

\begin{figure*}[!ht]
\centering 
\includegraphics[width=1.0\linewidth, height=0.12\linewidth]{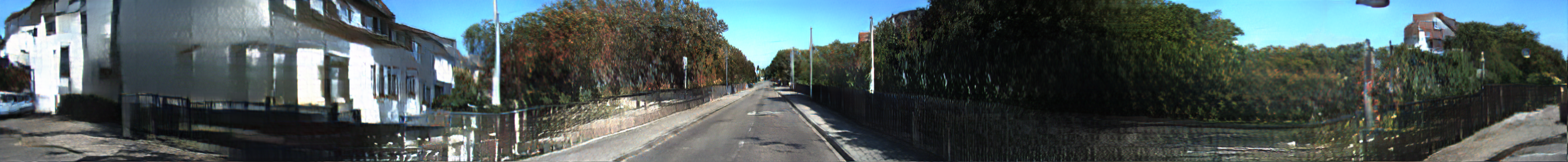}\par
\vspace{-0.05cm}
\includegraphics[width=1.0\linewidth, height=0.12\linewidth]{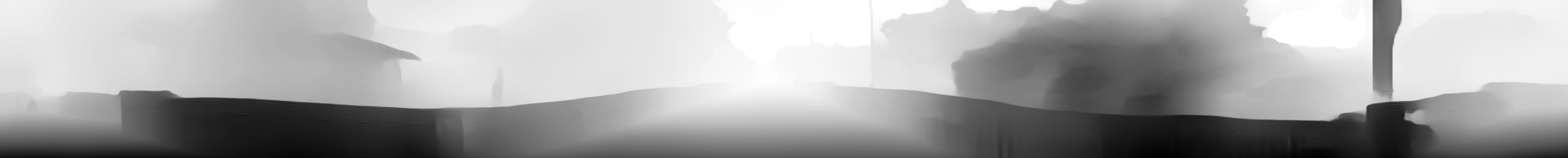}\par
\vspace{0.05cm}
\includegraphics[width=1.0\linewidth, height=0.12\linewidth]{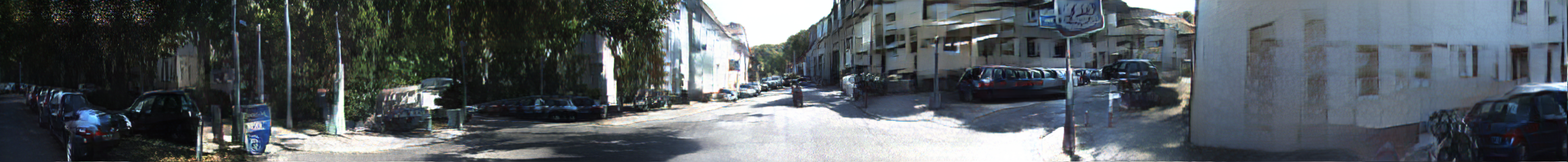}\par
\vspace{-0.05cm}
\includegraphics[width=1.0\linewidth, height=0.12\linewidth]{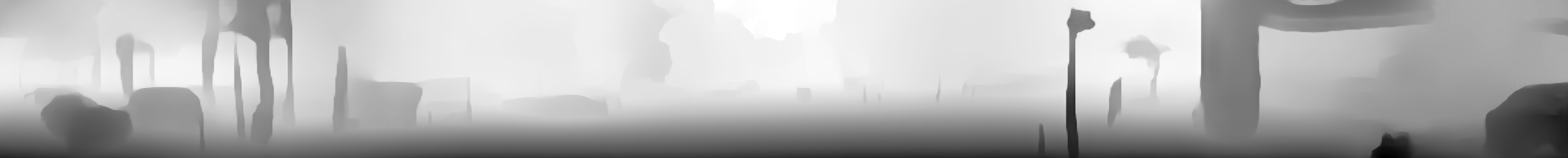}\par
\caption{Two sample panoramic images together with the corresponding depth estimates, synthesized by our proposed TITAN-Next model on the \sk test set.}
\label{fig:panoramic_images}
\end{figure*}

In contrast to other relevant works~\cite{Milz2019Points2Pix3P,Kim_2019,Kim2020ColorIG,Shalom2021}, our framework is the only work in the literature that has the capacity to process the full scan LiDAR data. 
This plays an important role to synthesize panoramic RGB-D scene images from the projected point clouds. Fig.~\ref{fig:panoramic_images} depicts two sample panoramic RGB images together with the corresponding depth maps, synthesized by our proposed TITAN-Next model on the \sk test split.
We here underline the fact that it is non-trivial to evaluate the quality of these generated panoramic RGB-D images, since there exists no corresponding ground truth data. 
The results reported so far in Figs.~\ref{fig:synt_img_results}-\ref{fig:wosemantic} show images synthesized only from a restricted region in the LiDAR projection, which approximates the original camera view as the only available ground truth.

%%%%%%%%%%%%%%%%%%%%%%%%%%%%%%%%%%%%%%%%%%%%%%%%%%%%%%%%%%%%%%%%%%%%%%%%%%%%%%%% 
 
%%%%%%%%%%%%%%%%%%%%%%%%%%%%%%%%%%%%%%%%%%%%%%%%%%%%%%%%%%%%%%%%%%%%%%%%%%%%%%%%

%%%%%%%%%%%%%%%%%%%%%%%%%%%%%%%%%%%%%%%%%%%%%%%%%%%%%%%%%%%%%%%%%%%%%%%%%%%%%%%%
%\begin{figure*}[!t]
%\centering 
%\includegraphics[width=1.0\linewidth, height=0.10\linewidth]{./images/qualitative/360/titannet/002685.png}\par
%\vspace{0.05cm}
%\includegraphics[width=1.0\linewidth, height=0.10\linewidth]{./images/qualitative/360/titannet++/002685.png}\par
%\vspace{0.2cm}
%\includegraphics[width=1.0\linewidth, height=0.10\linewidth]{./images/qualitative/360/titannet/001266.png}\par
%\vspace{0.05cm}
%\includegraphics[width=1.0\linewidth, height=0.10\linewidth]{./images/qualitative/360/titannet++/001266.png}\par
%\caption{Two sample panoramic images comparisons between TITAN-Net~\cite{Cortinhal_2021_ICCV} and our proposed TITAN-Next model on the \sk test set. The first and third image belong to TITAN-Net, while the second and fourth to TITAN-Next.}
%\label{fig:panoramic}
%\end{figure*}
%%%%%%%%%%%%%%%%%%%%%%%%%%%%%%%%%%%%%%%%%%%%%%%%%%%%%%%%%%%%%%%%%%%%%%%%%%%%%%%%
\begin{figure}[!t]
    \centering
    \subfigure{\includegraphics[width=0.5\textwidth]{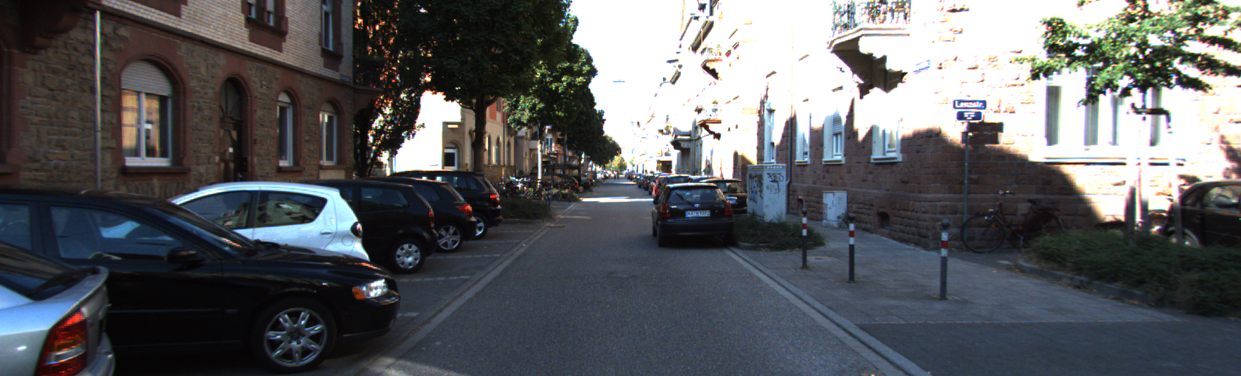}} 
    \subfigure{\includegraphics[width=0.5\textwidth]{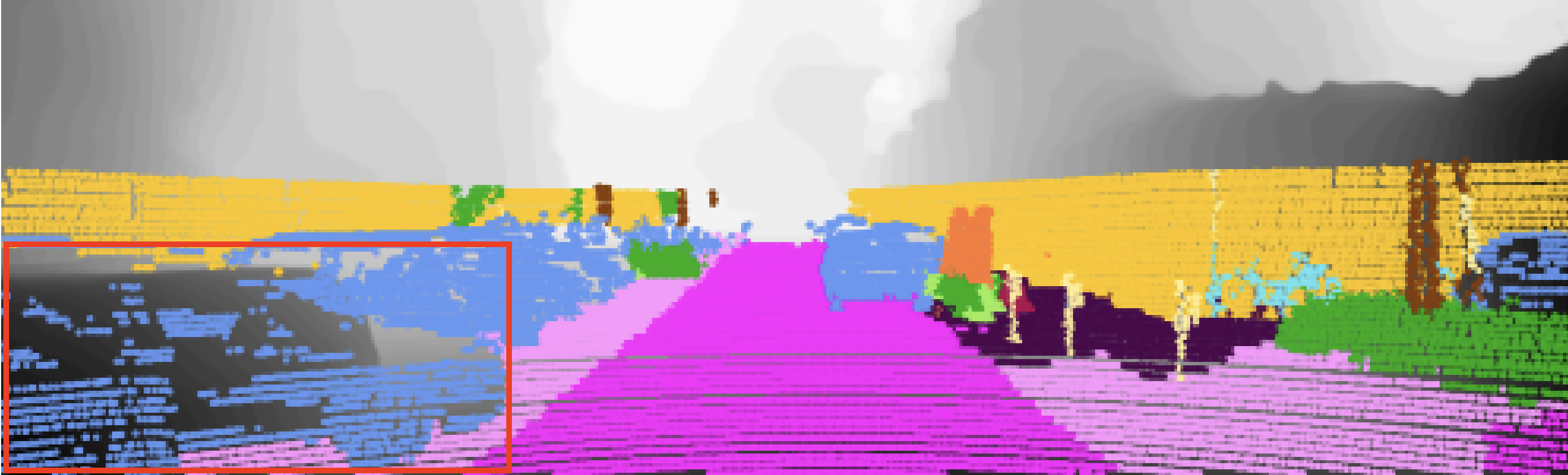}}
    \subfigure{\includegraphics[width=0.5\textwidth]{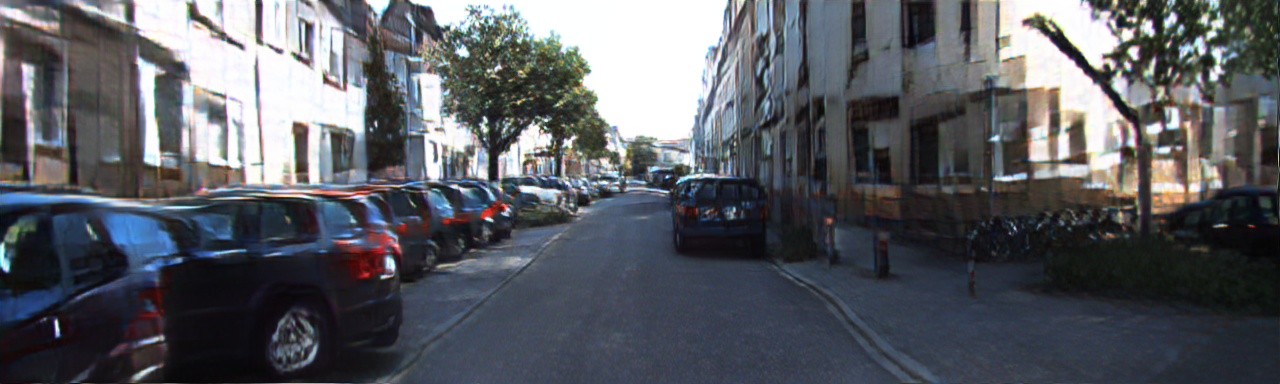}}
    \caption{A sample scene with a black car. From top to bottom, we have the original (ground-truth) RGB image, LiDAR segments superimposed on the predicted depth map, and the synthesized image, respectively. The red rectangle highlights the dense depth estimate for the region where the LiDAR data becomes sparse. }
    \label{fig:blackcar}
\end{figure}
%%%%%%%%%%%%%%%%%%%%%%%%%%%%%%%%%%%%%%%%%%%%%%%%%%%%%%%%%%%%%%%%%%%%%%%%%%%%%%%%

There exist potential applications of such depth- and semantics-aware domain translations, particularly in the field of autonomous vehicles. 
For instance, failed modality readings (e.g., camera images) can be substituted by the data predicted from other functional sensors (e.g., LiDAR) to feed the subsequent sensor fusion and planning modules. 
These predictions can form the initial beliefs about the scene to make intelligent decisions, instead of naively triggering an emergency stop.

In addition, this framework can also be employed to enrich the scene perception in the source domain, i.e., in the LiDAR space. 
For instance, visual object detection could be concurrently performed on photo-realistic  RGB-D image streams reconstructed from LiDAR point clouds. 
This way, black vehicles can be detected with higher accuracy, which is otherwise non-trivial to detect directly from LiDAR point clouds due to the low LiDAR reflectivity as shown in~\cite{Shalom2021}. 
Fig~\ref{fig:blackcar} shows a sample scene including a black vehicle. As highlighted by the red rectangle in this figure, the predicted depth map of the black car is quite dense, although the LiDAR point cloud becomes relatively sparse around the same vehicle.

Furthermore, such multimodal domain translation can help augment the already collected data, which is of utmost importance in regularizing the network training. 
By simply replacing the Vid2Vid head with models trained on other datasets, such as Cityscapes~\cite{Cityscapes}, a different variant of the same scene can be fabricated at no extra cost. 
Last but not least, this framework can also be employed to colorize point clouds without relying on camera readings. The work in~\cite{Vora_2020_CVPR} is one good application of such a point painting operation.

The main limitation of the proposed framework is the visual artifacts, such as broken vehicle boundaries, that emerge in the generated RGB images (see Fig.~\ref{fig:panoramic_images}). There are two main reasons for this. 
First, the semantic segmentation models (\snx and SD-Net) in our framework are not instance-aware approaches, thus, object boundaries between the same instances are not distinguishable. 
Second, our framework works frame-wise without incorporating the temporal coherence between adjacent frames. 
We believe that relying on individual segment instances while maintaining temporal consistency between frames can largely mitigate these visual artifacts in synthesized images. 
Thirdly, an observed misalignment arises between the semantic segmentation maps and depth estimations, attributed to the alignment approach adopted for bridging the LiDAR and camera perspectives. Our current methodology involves cropping the rangeview image to match the horizontal field of view of the camera. To enhance this alignment and foster improved correspondence between LiDAR data and depth estimations, further refinements to our technique are warranted.
%We refer the readers to the supplementary material providing more images and videos to show the TITAN-Next performance on the validation split.

We refer the readers to the supplementary videos\footnote{\href{https://youtu.be/rDmLPcpV2OM}{Panoramic Test Scene Video}}$^,$\footnote{\href{https://youtu.be/8VRDi5mB_sI}{Test Scene Video}}$^,$\footnote{\href{https://youtu.be/Ue6AMdns83Y}{Test Scene Video}}$^,$\footnote{\href{https://youtu.be/SE1qBrInvtg}{Test Scene Video}}$^,$\footnote{\href{https://youtu.be/TqwbGaoKCAo}{Depth Comparison Video}}
to show the TITAN-Next performance on the validation and test splits.
 
%We chose Vid2Vid-Net~\cite{wang2018vid2vid} as it performed the same task on a similar dataset. Nevertheless, a more suitable synthesizer could drastically improve RGB images, which - in turn - could be easily tested due to the modular pipeline structure.
%Generative approaches tend to introduce temporal flickering.

%%%%%%%%%%%%%%%%%%%%%%%%%%%%%%%%%%%%%%%%%%%%%%%%%%%%%%%%%%%%%%%%%%%%%%%%%%%%%%%% 

\section{Conclusion}

We present a new modular generative framework to address the multi-modal domain translation problem. More specifically, we introduce a depth- and semantics-aware conditional GAN model, named TITAN-Next which converts  LiDAR point cloud segments to the camera space to synthesize panoramic RGB-D camera images.
TITAN-Next builds up on the TITAN-Net~\cite{Cortinhal_2021_ICCV} and can achieve 23.7\% more accuracy for translating the semantic segments (see Table~\ref{tab:gen_sem_seg}). 
TITAN-Next differs in that it can construct the scene image enriched with semantic segments and depth information using the raw LiDAR point cloud only. 
Our work can be found on \href{https://github.com/TiagoCortinhal/TITAN-Next}{GitHub}.

%%%%%%%%%%%%%%%%%%%%%%%%%%%%%%%%%%%%%%%%%%%%%%%%%%%%%%%%%%%%%%%%%%%%%%%%%%%%%%%%

%%%%%%%%%%%%%%%%%%%%%%%%%%%%%%%%%%%%%%%%%%%%%%%%%%%%%%%%%%%%%%%%%%%%%%%%%%%%%%%%%%%%%%%%%%%%%%%%%%%%%%%%%%%%%
\section*{Acknowledgements}
This work was co-funded by the European Union (grant no. 101069576) and supported by Innovate UK (contract no. 10045139) and the Swiss State Secretariat for Education, Research and Innovation (contract no. 22.00123). Views and opinions expressed are however those of the author(s) only and do not necessarily reflect those of the European Union or the European Climate, Infrastructure and Environment Executive Agency (CINEA). Neither the European Union nor the granting authority can be held responsible for them.

%%%%%%%%%%%%%%%%%%%%%%%%%%%%%%%%%%%%%%%%%%%%%%%%%%%%%%%%%%%%%%%%%%%%%%%%%%%%%%%%%%%%%%%%%%%%%%%%%%%%%%%%%%%%%
%\section*{References}

% \bibliographystyle{plainnat}

{\footnotesize
\bibliographystyle{splncs04}
\bibliography{titanBib.bib}
}

\end{document}